\documentclass[conference]{IEEEtran}
\IEEEoverridecommandlockouts
\usepackage{cite}
\usepackage{amsmath,amssymb,amsfonts}
\usepackage{graphicx}
\usepackage{textcomp}
\usepackage{balance}

\usepackage{algorithm}
\usepackage{algorithmic}

\usepackage[table]{xcolor} 
\definecolor{heatmax}{HTML}{FDBE85} 
\definecolor{heatmin}{HTML}{FFF2CC} 
\definecolor{myBlue}{RGB}{68,114,196}
\definecolor{BrickRed}{RGB}{203,65,84}
\definecolor{ForestGreen}{RGB}{34,139,34}

\usepackage{booktabs}
\usepackage{multirow}
\usepackage{makecell}
\usepackage{float} 
\usepackage[font=footnotesize,labelfont=sc,labelsep=colon]{caption}
\usepackage[caption=false,font=footnotesize]{subfig}

\usepackage[most]{tcolorbox} 
\usepackage{enumitem}        
\usepackage[htt]{hyphenat}   

\usepackage{siunitx}
\usepackage[hidelinks]{hyperref}

\usepackage{etoolbox}
\makeatletter
\pretocmd{\@maketitle}{%
  \noindent\footnotesize\bfseries Currently under review\\[-0.25em]%
  \rule{\linewidth}{0.4pt}\par\vspace{0.8em}%
}{}{}
\makeatother

\begin{document}

\title{GradeSQL: Test-Time Inference with Outcome Reward Models for Text-to-SQL Generation from Large Language Models}

\author{
\IEEEauthorblockN{Mattia Tritto\textsuperscript{*}\textsubscript{A},
Giuseppe Farano\textsuperscript{*}\textsubscript{A},
Dario Di Palma\textsuperscript{*\dag}\textsubscript{A},
Gaetano Rossiello\textsubscript{B},\\
Fedelucio Narducci\textsubscript{A},
Dharmashankar Subramanian\textsubscript{B},
Tommaso Di Noia\textsubscript{A}}

\IEEEauthorblockA{\textsuperscript{A}\textit{Polytechnic University of Bari}, Bari, Italy\quad \textsuperscript{B}\textit{IBM T.J. Watson Research Center}, Yorktown Heights, NY, USA\\
\{mattiatritto01, farano.giuseppe02\}@gmail.com,
\{dario.dipalma, fedelucio.narducci, tommaso.dinoia\}@poliba.it, \\
gaetano.rossiello@ibm.com, dharmash@us.ibm.com}

\thanks{\textsuperscript{*}These authors contributed equally.}
\thanks{\textsuperscript{\dag}Corresponding author: dario.dipalma@poliba.it}
}

\maketitle

\begin{abstract}
Text-to-SQL, the task of translating natural language questions into SQL queries, has significantly advanced with the introduction of Large Language Models (LLMs), broadening database accessibility for a wide range of users. Despite substantial progress in generating valid SQL, current LLMs still struggle with complex queries. To address this limitation, \textit{test-time strategies} such as Best-of-$N$ (BoN) and Majority Voting (Maj) are often employed, based on the assumption that LLMs can produce correct answers after multiple attempts. However, these methods rely on surface-level heuristics, selecting the syntactically correct query through execution-based BoN (ex-BoN) or the most frequently generated one through Majority Voting. Recently, Outcome Reward Models (ORMs), which assign utility scores to generated outputs based on semantic correctness, have emerged as a promising reinforcement learning approach for improving model alignment. We argue that ORMs could serve as an effective new test-time heuristic, although their application in this context remains largely underexplored.

In this work, we propose a unified framework for training ORMs tailored to the Text-to-SQL task and assess their effectiveness as a test-time heuristic within the BoN strategy. We benchmark ORMs against ex-BoN and Maj across the BIRD and Spider datasets, fine-tuning diverse open-source LLMs from the Qwen2, Granite3, and Llama3 families. Results show that ORMs outperform ex-BoN and Maj, achieving execution accuracy gains of +4.33\% (BIRD) and +2.10\% (Spider) over ex-BoN, and +2.91\% (BIRD) and +0.93\% (Spider) over Maj. We further demonstrate that finetuning models already aligned with SQL generation, such as OmniSQL, yields superior ORM performance. Additionally, we observe that ORMs achieve competitive results on simple queries and benefit more from an increased number of candidates compared to ex-BoN and Maj. All code, datasets, and trained models are publicly released to support reproducibility and encourage future research in this area.
\end{abstract}

\begin{IEEEkeywords}
Large Language Models (LLMs), Text-to-SQL, Test-time Inference, Outcome Reward Models, Semantic Query Ranking.
\end{IEEEkeywords}

\section{Introduction}
Text-to-SQL, the task of translating Natural Language (NL) questions into executable SQL queries, facilitates the use of data management systems~\cite{DBLP:conf/iceis/NascimentoGFVIO24} and represents a step toward democratizing database access~\cite{DBLP:journals/pvldb/KimSHL20}, enabling both expert and non-expert users to interact with complex database systems through NL~\cite{DBLP:journals/nle/AndroutsopoulosRT95, DBLP:journals/pvldb/LiLCLT24, liu2025survey}.

Early approaches 
were predominantly rule-based, relying on manually crafted rules and heuristics to map NL questions to SQL queries~\cite{DBLP:journals/pvldb/LiJ14, mahmud2015rule, DBLP:conf/sigmod/LiJ14}. While often capable of generating syntactically correct queries, these systems struggled with challenges such as nested clauses, co-references, and complex aggregations~\cite{DBLP:journals/corr/abs-2406-08426}.

The field progressed with the adoption of neural network-based techniques~\cite{DBLP:conf/acl/XiaoDG16, DBLP:conf/acl/BoginBG19}, which improved intent recognition and semantic parsing. This was followed by the introduction of Pre-trained Language Models (PLMs) such as BERT~\cite{DBLP:conf/naacl/DevlinCLT19} and T5~\cite{DBLP:journals/jmlr/RaffelSRLNMZLL20}, which achieved competitive results across various benchmarks~\cite{DBLP:conf/aaai/Li00023, DBLP:conf/aaai/LiHCQ0HHDSL23}. Nevertheless, PLM-based systems still face difficulties with schema generalization and complex queries~\cite{liu2025survey}.

Recently, the advent of Large Language Models (LLMs) has introduced a new paradigm in Text-to-SQL. LLMs demonstrate strong emergent capabilities that surpass PLMs across diverse NLP tasks~\cite{DBLP:journals/air/Kumar24}, and they currently dominate research in this area~\cite{DBLP:conf/nips/PourrezaR23, DBLP:journals/pvldb/GaoWLSQDZ24, DBLP:journals/pacmmod/LiZLFZZWP0024, DBLP:conf/sigmod/ZhangMFMG0LL24}. Nevertheless, LLMs still face challenges in generating correct SQL for complex queries involving multi-table joins, nested subqueries, and sophisticated aggregations~\cite{DBLP:journals/corr/abs-2407-19517, DBLP:journals/elektrik/KanburogluT24}. To address these limitations, researchers are exploring techniques such as prompt engineering~\cite{DBLP:journals/pvldb/GaoWLSQDZ24}, multi-agent frameworks~\cite{DBLP:conf/coling/WangR0LBCYZYSL25, DBLP:conf/aaai/AskariPT25}, and fine-tuning~\cite{DBLP:journals/pacmmod/LiZLFZZWP0024, DBLP:conf/iclr/QinCF0P0Y25}. 
While these approaches have driven notable advances, achieving further improvements remains challenging due to the inherent complexity of the task, and even minor gains represent substantial progress~\cite{DBLP:journals/pvldb/GaoWLSQDZ24}.

Inspired by recent advances in mathematical reasoning with LLMs~\cite{DBLP:conf/acl/ZhangZWZLYLZL25}, the research community is increasingly investigating \textit{test-time inference} strategies, which allocate additional computational resources at inference to enhance output quality~\cite{DBLP:journals/pvldb/LiWZHZJWZCSCL25, DBLP:conf/acl/He0ZPWL25, DBLP:journals/corr/abs-2407-21787}. These strategies typically generate multiple candidate outputs and apply heuristics to select the best one, leveraging the ability of LLMs to produce correct answers when afforded more inference time rather than larger model size or additional training~\cite{DBLP:conf/acl/Xie0YZS25}. Common approaches include Best-of-$N$ (BoN)~\cite{DBLP:journals/corr/abs-2110-14168}, which samples $N$ outputs and selects the highest-scoring candidate, and Majority Voting (Maj)~\cite{DBLP:journals/corr/abs-2505-13271}, which chooses the most frequently generated output. However, these methods often depend on shallow heuristics such as frequency or execution success, without explicitly assessing semantic correctness.  

To address this gap, in this work we look beyond surface-level heuristics toward methods that explicitly assess the semantic quality of candidate outputs. In particular, the Reinforcement Learning (RL) community has proposed Outcome Reward Models (ORMs), which assign fine-grained semantic correctness scores to candidate outputs~\cite{DBLP:conf/acl/ZhangZWZLYLZL25}. When used as reward functions in RL loops, ORMs improve the alignment between model predictions and user intent~\cite{DBLP:journals/corr/abs-2403-04642}, highlighting their potential as principled verifiers of model output. This naturally raises the question: \emph{can ORMs also enhance test-time inference strategies?} In particular, integrating ORMs into BoN offers a way to move beyond frequency- or execution-based heuristics, enabling candidate selection grounded in semantic correctness. Although BoN and Maj have already been applied to Text-to-SQL~\cite{DBLP:journals/pvldb/LiWZHZJWZCSCL25}, the use of ORMs in test-time inference remains largely unexplored, hindered by the scarcity of reward-labeled data and the absence of clear methodological guidelines. Moreover, no systematic evaluation has yet compared ORM-guided selection with traditional test-time compute strategies, leaving open the potential of ORMs as a more robust heuristic for BoN.


To address these limitations, we first introduce a scalable data synthesis framework tailored for Outcome Reward Models in the Text-to-SQL setting, which automates data generation to mitigate data scarcity and enables the development of task-specific ORMs. Building on this foundation, we present a comprehensive evaluation of ORM-guided Best-of-$N$, benchmarking it against execution-based BoN and Majority Voting.

Specifically, the framework consists of three stages: \textit{candidate generation}, \textit{data labeling}, and \textit{supervised fine-tuning}. First, we employ a powerful LLM to generate a diverse set of SQL query candidates for a given natural language question and database schema. This set includes both correct queries, which yield execution results consistent with the gold standard, and incorrect but syntactically valid alternatives. Next, we label these candidates to construct a dataset of positive and negative examples. Finally, this dataset is used to fine-tune a separate LLM, which serves as an ORM. The trained ORM learns to score candidate SQL queries based on their alignment with the original user intent and the database schema. At inference, the ORM functions as a post-generation re-ranking module, selecting the most semantically faithful query from the candidate pool generated by the base model.


To evaluate the effectiveness of our approach, we conduct a comprehensive study using  open-source LLMs in an end-to-end Text-to-SQL setup.



Our experiments yield several key insights into the effectiveness of ORMs for Text-to-SQL: (1) ORMs consistently outperform both execution-based Best-of-$N$ and Majority Voting, achieving execution accuracy gains of up to +4.33\% on BIRD and +2.10\% on Spider over ex-BoN, and +2.91\% (BIRD) and +0.93\% (Spider) over Maj; (2) fine-tuning ORM models that are already aligned with SQL generation tasks, such as OmniSQL, further enhances performance; (3) ORMs perform competitively on simple queries while demonstrating greater robustness on complex ones compared to ex-BoN and Maj; and (4) ORM performance benefits more substantially from increasing the number of candidate queries than the baseline methods.

\noindent In summary, our work makes the following contributions:
\begin{itemize}[itemsep=0pt, parsep=0pt, topsep=0pt, partopsep=0pt, leftmargin=15pt]
    \item \textbf{Scalable Data Synthesis and ORM Training:} We introduce a framework to synthesize high-quality annotated datasets for ORM fine-tuning, overcoming data scarcity in Text-to-SQL reward modeling.
    \item \textbf{Extensive Empirical Evaluation and Analysis:} We conduct a large-scale study across LLM families and parameter sizes, analyze test-time efficiency under candidate scaling, and provide controlled ablations on prompt design, model scale, and training losses.
    \item \textbf{Open-Source Resources:} We release our codebase, synthesized datasets, and fine-tuned ORMs on GitHub and Hugging Face to promote reproducibility and future research.
\end{itemize}

\section{Related Work}
In this section, we trace the evolution of Text-to-SQL solutions, review test-time inference strategies, and conclude with an overview of existing data-augmentation frameworks for Text-to-SQL.

\subsection{Text-to-SQL Meets LLMs}
Text-to-SQL refers to the task of translating Natural Language (NL) questions into executable SQL queries, serving as a bridge between human intent and structured data access. Its development spans more than five decades, evolving from early handcrafted systems to today’s approaches powered by Large Language Models (LLMs)~\cite{DBLP:journals/corr/abs-2406-08426}.  

The earliest systems were entirely rule-based, relying on domain-specific grammars and manually engineered parsing rules. A seminal example is LUNAR (1971)~\cite{woods1972lunar}, which answered questions about lunar rock samples using an Augmented Transition Network parser combined with procedural semantics. In the 1980s, the field began adopting intermediate logic-based representations, where systems first converted NL into a database-agnostic logical form before generating SQL. The logic-based interpreter by Warren and Pereira (1982)~\cite{DBLP:journals/coling/WarrenP82} exemplifies this trend. By the late 1990s and early 2000s, a shift toward data-driven methods had emerged, as researchers explored whether parsers could be trained statistically rather than handcrafted~\cite{DBLP:conf/emnlp/TangM00}.  

The mid-2010s marked a paradigm shift with the advent of deep learning. Neural semantic parsers were capable of jointly modeling questions and database schemas, but progress was constrained by the lack of large-scale labeled datasets~\cite{DBLP:journals/pvldb/KimSHL20}. A breakthrough came with the release of
Spider (2018)~\cite{DBLP:conf/emnlp/YuZYYWLMLYRZR18}, which provided benchmarks for cross-domain generalization. Building on these datasets, Pre-trained Language Models (PLMs) such as BERT~\cite{DBLP:conf/naacl/DevlinCLT19} and T5~\cite{DBLP:journals/jmlr/RaffelSRLNMZLL20} catalyzed the development of schema-aware neural parsers. RAT-SQL~\cite{DBLP:conf/acl/WangSLPR20}, for example, integrated relation-aware transformers with pre-trained embeddings to significantly improve schema linking, i.e., mapping question terms to database elements.

The advent of LLMs such as GPT~\cite{DBLP:journals/corr/abs-2303-08774}, Llama~\cite{DBLP:journals/corr/abs-2407-21783}, and PaLM~\cite{DBLP:journals/corr/abs-2305-10403} has further transformed the landscape. Unlike PLMs, LLMs exhibit strong emergent capabilities, enabling zero- and few-shot generalization with minimal or no task-specific fine-tuning. Studies such as those by Liu et al.~\cite{DBLP:journals/corr/abs-2303-13547} have shown that ChatGPT achieves up to 70\% execution accuracy on Spider with only prompt-based methods. Few-shot prompting strategies, such as DIN-SQL~\cite{DBLP:conf/nips/PourrezaR23}, decompose the task into subtasks to improve accuracy, though performance still degrades on highly complex queries. More recent work combines LLMs with Retrieval-Augmented Generation (RAG)~\cite{DBLP:conf/is/VichevM24,DBLP:journals/corr/abs-2501-17174}, which injects schema- or database-specific context at inference time, yielding improved accuracy on large, content-rich benchmarks such as BIRD (2023)~\cite{DBLP:conf/nips/LiHQYLLWQGHZ0LC23}.

Building on these advances, recent research increasingly frames Text-to-SQL as a reasoning problem. Rather than treating query generation as a single-shot task, approaches leverage LLMs for multi-step decomposition, error correction, and domain adaptation. Pipelines often decompose the task into components such as schema linking, query generation, refinement, and selection, orchestrated through prompt engineering or fine-tuning~\cite{DBLP:conf/acl/He0ZPWL25, DBLP:conf/acl/HongYCZH024}. Some frameworks adopt multi-agent designs~\cite{DBLP:conf/aaai/AskariPT25}, while others focus on scaling SQL-specific training data~\cite{DBLP:journals/pvldb/LiWZHZJWZCSCL25}. Current trends emphasize expanding reasoning capacity~\cite{DBLP:journals/pvldb/GaoWLSQDZ24,DBLP:journals/corr/abs-2502-06759}, integrating domain knowledge~\cite{DBLP:conf/nips/LiHQYLLWQGHZ0LC23}, and refining fine-tuning strategies~\cite{DBLP:journals/pacmmod/LiZLFZZWP0024}.  

Despite this rapid progress, the intrinsic complexity of Text-to-SQL continues to limit accuracy, especially on queries requiring multi-table joins, nested subqueries, and sophisticated aggregations. These challenges have sparked growing interest in \emph{test-time inference} strategies, which allocate additional computational resources at inference time to enhance output quality~\cite{DBLP:journals/pvldb/LiWZHZJWZCSCL25, DBLP:conf/acl/He0ZPWL25, DBLP:journals/corr/abs-2407-21787}.

\subsection{Test-Time Inference Strategies}
Test-Time Inference (TTI), also known as inference-time scaling or test-time computation, refers to strategies that improve LLM outputs by allocating additional computation during inference without modifying model parameters~\cite{DBLP:conf/iclr/Snell0XK25}. Nowadays, LLMs generate text by sequentially predicting tokens, with decoding strategies that determine which token to produce at each step. Early approaches relied on deterministic methods such as greedy decoding~\cite{DBLP:journals/jmlr/Chickering02a} or beam search~\cite{DBLP:journals/coling/TillmannN03}, which maximized sequence likelihood but often produced repetitive outputs with limited diversity~\cite{DBLP:journals/tacl/MeisterWC22}. To address this, later research introduced stochastic decoding algorithms, including temperature scaling, top-$k$ sampling~\cite{DBLP:conf/acl/LewisDF18}, and nucleus (top-$p$) sampling~\cite{DBLP:conf/iclr/HoltzmanBDFC20}, which add controlled randomness by sampling from truncated distributions. Although effective in improving local variability, these methods operate only at the token level and thus cannot substantially improve the correctness of full outputs. TTI builds on these foundations by shifting the focus from token-level diversity toward strategies that explicitly leverage multiple candidate generations, verification mechanisms, and structured reasoning to enhance robustness and accuracy.  

Building on this idea, commonly adopted methods such as Best-of-$N$ (BoN) and Majority Voting (Maj) have emerged~\cite{DBLP:conf/iclr/Snell0XK25}. BoN selects the candidate with the highest score according to a heuristic function, such as model likelihood~\cite{DBLP:conf/iclr/HoltzmanBDFC20} or execution accuracy~\cite{DBLP:conf/iclr/ChenZNZLLC23}, while Maj chooses the most frequent final answer among diverse samples~\cite{DBLP:conf/iclr/0002WSLCNCZ23}. These approaches are attractive because they require no retraining, but they also inherit critical limitations: BoN tends to over-prefer generic high-probability outputs~\cite{DBLP:journals/tacl/MeisterPWC23, DBLP:conf/acl/LiHFLEHZL23}, while Maj can fail when correct solutions are underrepresented~\cite{DBLP:conf/iclr/0002WSLCNCZ23, DBLP:conf/emnlp/0003MDMWL24}. This motivated the development of verification-guided selection methods, which go beyond surface heuristics by introducing an explicit verifier to score or filter candidates.  

A major advancement in this direction came with reward models. Process Reward Models (PRMs)~\cite{DBLP:journals/corr/abs-2504-16828} evaluate intermediate reasoning steps but require costly step-level supervision. In contrast, Outcome Reward Models (ORMs)~\cite{DBLP:journals/corr/abs-2110-14168} score only final outputs, avoiding intermediate annotation and achieving notable success in domains such as mathematical reasoning~\cite{DBLP:journals/corr/abs-2110-14168}. In Text-to-SQL, ORMs are particularly appealing because they can semantically evaluate SQL queries independently of the generator. This enables the selection of high-quality but underrepresented candidates~\cite{DBLP:journals/corr/abs-2503-24364, DBLP:conf/acl/He0ZPWL25}, a capability missing in Maj, while also identifying semantically equivalent but syntactically different queries. These properties make ORMs a natural fit for re-ranking diverse candidate sets and for modular system designs where generation and verification can be optimized independently.  

\textbf{Datasets for ORM Training.}  
A central challenge in realizing ORM potential is the scarcity of large, high-quality datasets that pair inputs with multiple candidate outputs and quality labels. In Text-to-SQL, most dataset construction efforts have focused on improving general model training rather than verifier-specific needs. Early approaches used SQL-to-question pipelines, where SQL queries generated from templates were translated into natural language~\cite{DBLP:conf/emnlp/GuoSTDYCCCZ18, DBLP:conf/naacl/KobayashiLSCGZWN25, DBLP:journals/pvldb/ZhangDKKKS23, DBLP:conf/emnlp/ZhongLWZ20}. While effective, these methods were labor intensive and limited in diversity. More recent LLM-based approaches, such as SENSE~\cite{DBLP:conf/acl/YangHY0LZ24}, improve realism through multi-step generation but rely on expensive proprietary models. Syn-SQL~\cite{DBLP:journals/pvldb/LiWZHZJWZCSCL25} mitigates this by decomposing synthesis into smaller steps that can be executed with open-source LLMs, producing 2.5M examples in a scalable and reproducible manner.  

Despite these advances, no prior work has directly tackled the unique requirements of ORM training, where datasets must support fine-grained verification across diverse candidate outputs rather than solely boosting general SQL generation performance. This gap underscores the need for new methodologies that explicitly target ORM development.

\section{Methodology}
\subsection{Problem Formulation.}
Given a natural language question $q_i \in Q$, where $Q$ denotes the set of all questions, the Text-to-SQL task aims to map $q_i$ and its corresponding database schema $\Sigma_i$ (a collection of tables with their columns) into a valid SQL query $y_{gold_i} \in Y_{\text{gold}}$~\cite{DBLP:conf/acl/BaekSHSSGB25, DBLP:conf/acl/DaiYMC25}.

While the traditional formulation focuses on generating a single query, in the context of \emph{test-time inference} the process is extended to consider multiple candidates. Specifically, given $q_i$, an LLM generates a pool of $N$ candidate queries
\[
Y_{\text{candidate}} = \{c_1, c_2, \dots, c_N\},
\] 
from which either a heuristic or a verifier must identify the candidate $c_j \in Y_{\text{candidate}}$ that best corresponds to the ground-truth gold query. In this work, we define the verifier as an \emph{Outcome Reward Model (ORM)}.  

Formally, the ORM is a scoring function $\phi(q_i, c_j)$ that maps a natural language question $q_i$ and a candidate SQL query $c_j \in Y_{\text{candidate}}$ to a probabilistic score:
\[
\phi(q_i, c_j) \in [0, 1],
\]
where higher scores indicate a greater likelihood that the query is correct. The ORM thus induces a ranked list of candidate SQL queries, with the best candidate $c^*$ selected as the one achieving the highest score:
\[
c^* = \operatorname*{arg\,max}_{c_j \in Y_{\text{candidate}}} \phi(q_i, c_j).
\]

Training an ORM as a scoring function for candidate ranking introduces two central challenges:  
(i) constructing datasets that explicitly identify which candidate $c_j$ is correct, and  
(ii) assigning relative scores to multiple candidates in a way that faithfully reflects user intent.  

To overcome these challenges, we leverage the LLM’s internal ability to self-assess~\cite{DBLP:conf/naacl/LiuLZDCHXCW24, DBLP:conf/emnlp/HuangQLYXZL24, DBLP:conf/emnlp/LiWFZWC24} query correctness. Specifically, we prompt the model to classify each candidate as ``\texttt{Yes}'' (correct) or ``\texttt{No}'' (incorrect), and then extract the associated logits. These logits are used during training to supervise the ORM, enabling it to learn a probabilistic scoring function that generalizes to ranking candidates at inference time.  

In the following, we present the \textbf{GradeSQL} framework, which implements this idea by constructing and training ORMs specifically designed for test-time inference in the Text-to-SQL setting.

\begin{figure*}[t]
    \centering
    \resizebox{1.0\textwidth}{!}{%
        \includegraphics{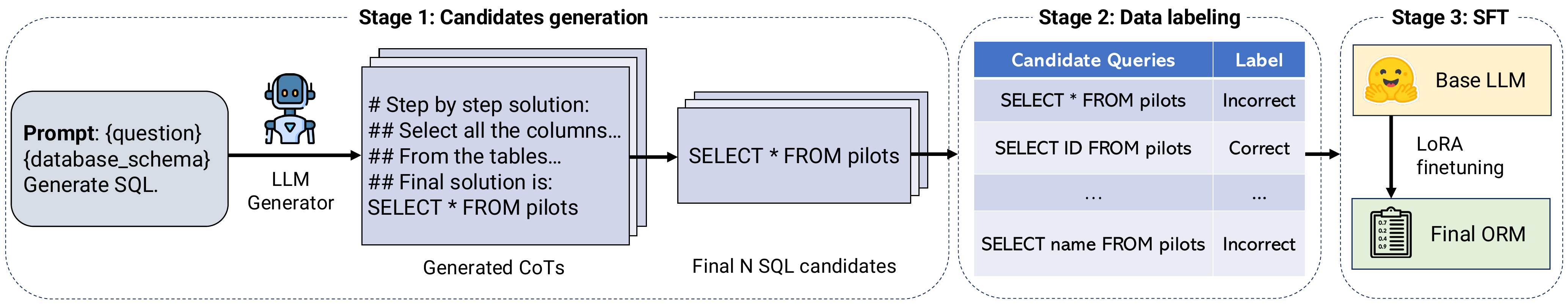}
    }
    \caption{Overview of the \textit{GradeSQL} framework for training an ORM. The framework consists of three stages: (i) \textbf{Candidate Generation}, an LLM generates $N$ SQL queries via a Chain-of-Thought (CoT) prompt, then removes the CoT trace; (ii) \textbf{Data Labeling}, each candidate is annotated as correct or incorrect; and (iii) \textbf{Supervised Fine-Tuning (SFT)}, the base LLM is fine-tuned with LoRA on the labeled data to obtain the final ORM.}
    \label{fig:orm_framework}
\end{figure*}

\subsection{GradeSQL Framework}
Given a Text-to-SQL dataset \(\mathcal{D} = \{(q_i, y_{gold_i})\}_{i=1}^n\), where \(q_i\) is a natural language question and \(y_{gold_i}\) its corresponding gold SQL query, our framework (Figure~\ref{fig:orm_framework}) comprises three stages: (i) Candidates generation, (ii) Data labeling, and (iii) Supervised Fine-Tuning (SFT). The overall objective is to train an Outcome Reward Model (ORM) tailored to the target dataset.  



\noindent\textbf{Stage 1: Candidate Generation.} For each question $q_i$, we employ an LLM, referred to as the generator $\mathcal{G}$, to produce a set of $N$ candidate SQL queries (the prompt is provided in Appendix~\ref{prompt:SQL_candidate_generation_prompt}).

Formally, let $q_i$ denote a natural language question and $\Sigma_i$ the corresponding database schema. The generator $\mathcal{G}(q_i,\Sigma_i)$ produces $N$ candidate SQL queries:
\[
Y_{\text{candidate}} = \{\texttt{SQL}_1, \texttt{SQL}_2, \dots, \texttt{SQL}_N\} = \{c_1, c_2, \dots, c_N\}.
\]

\noindent\textbf{Stage 2: Data Labeling.} Once candidates are generated, the next step is to determine which of them correctly capture the intent of $q_i$.  
Let $R(c_j)$ denote the execution result of a query $c_j$, i.e., the set of tuples returned when $c_j$ is executed on the target database.  
A candidate $c_j \in Y_{\text{candidate}}$ is labeled as \textit{correct} if its result set matches that of the gold query $y_{gold_i}$, i.e., $R(c_j) = R(y_{gold_i})$.

Candidates are labeled as \textit{incorrect} if they fail to capture the intended semantics. This may occur in two cases:
\begin{enumerate}[label=(\arabic*), itemsep=1pt, parsep=1pt, topsep=1pt, partopsep=1pt, leftmargin=30pt]
    \item \underline{Semantic mismatch}: The query executes successfully but returns a different result set from the gold query, i.e., $R(c_j) \neq R(y_{gold_i})$.
    \item \underline{Execution error}: The query contains syntactic or runtime errors that cause the database management system (DBMS) to fail during execution.
\end{enumerate}

Since execution errors are trivial to detect, the framework primarily focuses on distinguishing semantically correct queries from incorrect but executable ones. This process is formalized by the labeling function $\ell(c_i)$:
\[
\ell(c_j) = 
\begin{cases}
\text{correct (Yes)} & \text{if } R(c_j) = R(y_{gold_i}), \\
\text{incorrect (No)} & \text{if } R(c_j) \neq R(y_{gold_i}), \\
\text{discarded} & \text{if } c_j \text{ raises an execution error}.
\end{cases}
\]

\noindent\textbf{Stage 3: Supervised Fine-tuning.} The labeled candidates form a training set $\mathcal{D}_{\text{label}} = \{(c_i, l_i)\}_{i=1}^n$, where $c_i$ is a candidate SQL query and $l_i \in \{\texttt{Yes}, \texttt{No}\}$ indicates correctness. At this stage, the verification task is cast as an autoregressive binary classification problem.  

Formally, let $x = \mathrm{Prompt}(\Sigma, q, c)$ denote the input sequence constructed from the database schema $\Sigma$, the natural language question $q$, and the candidate SQL query $c$ (see Appendix~\ref{prompt:sql-only_orm}). The label $l \in \{\texttt{Yes}, \texttt{No}\}$ specifies whether $c$ semantically satisfies $q$ under $\Sigma$. The model is trained to autoregressively generate the next token: ``\texttt{Yes}'' if correct and ``\texttt{No}'' otherwise.

We fine-tune the causal LM head as a verifier using Low-Rank Adaptation (LoRA)~\cite{DBLP:conf/iclr/HuSWALWWC22} under the causal language modeling objective~\cite{radford2019language}.  
Given the input $x = \mathrm{Prompt}(\Sigma, q, c)$, we append the label $l \in \{\texttt{Yes}, \texttt{No}\}$ to obtain the target sequence $s = [\,x;\, l\,]$.  
During training, the model conditions on $s$ and is optimized to minimize the negative log-likelihood:
\[
\mathcal{L}(\theta) \;=\; - \sum_{t=1}^{|s|} \log P_\theta\!\left(s_t \mid s_{<t}\right).
\]

This objective guides the model, when presented with a structured input $x = \mathrm{Prompt}(\Sigma, q, c)$, to autoregressively generate the correct label token $l \in \{\texttt{Yes}, \texttt{No}\}$.



\subsection{ORM Inference and Probabilistic Scoring Mechanism}
At inference time, the verifier assigns a probabilistic score to each candidate SQL query, reflecting its likelihood of being semantically correct. This is achieved by prompting the fine-tuned LLM in an autoregressive setting to generate a single verification token, either ``\texttt{Yes}'' or ``\texttt{No}''. Rather than treating this prediction as a discrete token-level decision, we instead use the model’s logit corresponding to the ``\texttt{Yes}'' token, normalized into $[0,1]$, as a continuous confidence score. In this way, the ORM functions not merely as a classifier but as a ranking mechanism over candidate queries. A visual representation of the inference process is provided in the Appendix (see Figure~\ref{fig:inference_schema}).

\paragraph{Autoregressive Verification.}  
Formally, for a given prompt $x = \mathrm{Prompt}(\Sigma, q, c)$, constructed from the schema $\Sigma$, natural language question $q$, and candidate  query $c$, the model defines a conditional distribution:
\[
P(y \mid x; \theta), \quad y \in \{\texttt{Yes}, \texttt{No}\},
\]
where $\theta$ are the model parameters. The verifier’s score for candidate $c$ is then:
\[
\text{Score}(c) = P(y_{\text{yes}} \mid x; \theta).
\]
This formulation avoids hard decision boundaries, instead producing a real-valued confidence score that can be directly used to rank candidates.  

\paragraph{Motivation for Probabilistic Scoring.}  
Framing ORM inference in terms of probabilities provides two key advantages for Text-to-SQL verification:  
\begin{enumerate}[label=(\arabic*), itemsep=1pt, parsep=1pt, topsep=1pt, partopsep=1pt, leftmargin=30pt]
    \item \underline{Fine-grained calibration:} Scores capture nuanced differences in semantic plausibility, even among syntactically valid queries.  
    \item \underline{Differentiable ranking:} The continuous nature of the scores enables consistent ordering of candidates, avoiding the brittleness of binary thresholding. 
\end{enumerate}

By interpreting the verifier as a probabilistic scoring function rather than a binary classifier, the ORM offers a mechanism for ranking candidate SQL queries by their semantic correctness.

\section{Experimental Setup}
\textbf{Datasets.} To evaluate our approach, we use two widely adopted Text-to-SQL benchmarks: Spider~\cite{DBLP:conf/emnlp/YuZYYWLMLYRZR18} and BIRD~\cite{DBLP:conf/nips/LiHQYLLWQGHZ0LC23}. Both are cross-domain benchmarks with disjoint databases across train/dev/test splits to assess generalization. On Spider, we construct the labeled dataset from the training questions, train the ORM, and evaluate it on the dev and test sets. On BIRD, we adopt the same setup, using the training questions to build the dataset and train the ORM, but, due to the hidden test set, evaluate only on the dev set.

\noindent \textbf{Metrics.} To ensure fair comparison across methods, we adopt the official evaluation metrics defined for BIRD~\cite{DBLP:conf/nips/LiHQYLLWQGHZ0LC23}, which are also applicable to Spider. Let $Q_{\text{test}}$ denote the set of evaluation questions. For each question $q_i \in Q_{\text{test}}$, let $c^*_i$ be the predicted SQL query and $y_{gold_i}$ the corresponding gold query. 

Execution Accuracy (EX)~\cite{DBLP:conf/nips/LiHQYLLWQGHZ0LC23} measures the proportion of questions for which the predicted and gold queries return identical result sets:
\[
\text{EX} = \frac{1}{|Q_{\text{test}}|} \sum_{q_i \in Q_{\text{test}}} \mathbf{1}\!\left(R(c^*_i) = R(y_{gold_i})\right),
\]
where $\mathbf{1}(\cdot)$ is the indicator function, returning 1 if the condition holds and 0 otherwise.

Pass@N~\cite{DBLP:journals/corr/abs-2107-03374} evaluates the proportion of questions for which at least one of the $N$ predicted candidates returns the correct result set:
\[
\text{Pass@N} = \frac{1}{|Q_{\text{test}}|} \sum_{q_i \in Q_{\text{test}}} \mathbf{1}\!\left(\bigvee_{j=1}^N R(c_j) = R(y_{gold_i})\right),
\]
where $\bigvee$ denotes the logical OR across $N$ candidates. Specifically, the prediction is considered correct if any of the $N$ candidates for question $q_i$ yield the same result set as the ground truth $y_{gold_i}$.

\noindent \textbf{Baselines.} We benchmark the ORM-based test-time inference (TTI) against two TTI strategies widely used in Text-to-SQL and code generation tasks: Majority Voting and execution-based Best-of-$N$. Both methods select a single SQL query from the generated pool without relying on gold references, but differ in how they aggregate candidates.

\paragraph{Majority Voting} Majority Voting selects the most plausible SQL query based on execution equivalence~\cite{DBLP:journals/corr/abs-2505-13271}. Each candidate query is executed on the target database, and queries producing identical result sets are grouped into $k$ clusters. The class with the largest cardinality is selected, and the final query is sampled uniformly from it. Formally, let $Y_{\text{candidate}} = \{c_1, c_2, \dots, c_N\}$ be the set of candidates.
We define
\[
c_i \sim c_j \;\;\text{if and only if}\;\; R(c_i) = R(c_j),
\]
which partitions $Y_{\text{candidate}}$ into $k$ clusters $C_1, \dots, C_k$. Majority Voting then selects the cluster with the largest cardinality and samples one candidate from it. This method is robust to surface-level variations and has been widely adopted in the literature, including BIRD~\cite{DBLP:conf/nips/LiHQYLLWQGHZ0LC23}.

\paragraph{\texorpdfstring{Execution-based Best-of-$N$}{Execution-based Best-of-$N$}}  
Best-of-$N$ selects the most promising candidate based on an execution-level heuristic~\cite{DBLP:conf/iclr/ChenZNZLLC23}. Given $Y_{\text{candidate}} = \{c_1, \dots, c_N\}$, the heuristic $h:Y_{\text{candidate}} \rightarrow [0,1]$ considers two signals: (i) \emph{executability} (whether the query runs without errors), and (ii) \emph{non-emptiness} (whether it returns at least one row). The scoring function is:
\[
h(c_i) =
\begin{cases}
0   & \text{if } c_i \text{ is not executable}, \\
0.5 & \text{if } c_i \text{ executes but returns 0 rows}, \\
1   & \text{if } c_i \text{ executes and returns at least one row}.
\end{cases}
\]
The final query $c^*$ is selected uniformly from the subset of candidates that achieve the highest score.

While this efficiently filters invalid queries, it provides only coarse discrimination: multiple queries may achieve the same maximal score while differing in semantic correctness. This limitation motivates our ORM-based extension of Best-of-$N$.

\begin{table}[t!]
\centering
\caption{Reproducibility results of OmniSQL-7B. Maj = Majority Voting (temperature $T{=}0.8$, $N{=}8$); Gre = Greedy decoding ($N{=}1$). The table compares our reproduction (Our) with the original results (Origin) on the BIRD dev, Spider dev, and Spider test benchmarks.}
\label{tab:omnisql_reproducibility}
\footnotesize
\renewcommand{\arraystretch}{1.08}
\setlength{\tabcolsep}{6pt}

\resizebox{\columnwidth}{!}{%
    \begin{tabular}{lcccccc}
    \toprule
     & \multicolumn{2}{c}{\textbf{BIRD dev}} & \multicolumn{2}{c}{\textbf{Spider dev}} & \multicolumn{2}{c}{\textbf{Spider test}} \\
    \cmidrule(lr){2-3} \cmidrule(lr){4-5} \cmidrule(lr){6-7}
     & Maj & Gre & Maj & Gre & Maj & Gre \\
    \midrule
    \textbf{Our}    & 66.95 & 64.41 & 83.95 & 82.11 & 85.61 & 84.49 \\
    \textbf{Origin} & 66.10 & 63.90 & 81.60 & 81.20 & 88.90 & 87.90 \\
    \bottomrule
    \end{tabular}
}

\end{table}
\noindent \textbf{ORMs and Evaluation Setup.}  
To evaluate ORM-based Best-of-$N$, we adopt OmniSQL-7B~\cite{DBLP:journals/pvldb/LiWZHZJWZCSCL25} as the generator, which, at the time of writing, is the most capable open-source 7B model trained specifically for the Text-to-SQL task. For each input question, OmniSQL produces $N$ candidate queries, which are then re-ranked by three strategies: Majority Voting, execution-based Best-of-$N$, and ORM-based Best-of-$N$. 

Before integration, we validated OmniSQL through a reproducibility study on Spider and BIRD. As shown in Table~\ref{tab:omnisql_reproducibility}, our reproduced results closely match those of the original work, with deviations of at most $1\%$ on BIRD dev and within $+3\%$ on Spider dev. This confirms the robustness of OmniSQL and justifies its use as our base generator, particularly in combination with ORM re-ranking.  


\noindent \textbf{Reproducibility details.} 
All experiments were conducted on compute nodes with Intel Xeon 8358 CPUs (32 cores, 2.6 GHz), 512 GB RAM, and 4 NVIDIA A100 SXM4 GPUs (64 GB). SQL candidates were generated via \texttt{vLLM} using the OmniSQL setup~\cite{DBLP:journals/pvldb/LiWZHZJWZCSCL25}, with ORM inference through Hugging Face \texttt{transformers}. Runs used a fixed seed of 42. Code\footnote{\href{https://github.com/sisinflab/GradeSQL}{GradeSQL Framework}}, datasets\footnote{\href{https://huggingface.co/collections/sisinflab-ai/gradesql-training-datasets-68ac62e1356b5399ef81236c}{GradeSQL Training Datasets}}, and models\footnote{\href{https://huggingface.co/collections/sisinflab-ai/gradesql-models-68ac58755ffe5fe880e0acb5}{Pretrained GradeSQL ORMs}} are publicly available.

\section{Results and Discussion}

\begin{table}[t]
\centering
\caption{Dataset statistics for \textit{BIRD} and \textit{Spider} (training splits). We report (i) the number of questions before and after de-duplication (with removals $\Delta$), and (ii) the final size of the labeled sets generated by sampling $N\!=\!32$ candidates per question. Percentages indicate the class distribution \emph{(Incorrect / Correct)} in the \emph{imbalanced} set; the balanced set is 50/50.}
\label{tab:dataset_stats}

\scriptsize
\setlength{\tabcolsep}{6pt}

\resizebox{\columnwidth}{!}{%
\begin{tabular}{@{}lrr@{}}
  \toprule
  & \textbf{BIRD (train)} & \textbf{Spider (train)} \\
  \midrule
  \multicolumn{3}{@{}l}{\textit{Questions}} \\
  \quad \# (pre)                 & 9,428   & 9,000   \\
  \quad \# (post)                & 9,411   & 8,960   \\
  \quad Removed $\Delta$ (\%)    & 17 (0.18\%) & 40 (0.44\%) \\
  \addlinespace[1pt]
  \multicolumn{3}{@{}l}{\textit{Labeled sets}} \\
  \quad Imbalanced size          & 82,640  & 50,073  \\
  \quad \,\,\,\,Class split (\% inc / cor)
                                 & 41.37 / 58.63
                                 & 30.60 / 69.40 \\
  \quad Balanced size (50/50)    & 30,686  & 17,834  \\
  \bottomrule
\end{tabular}%
}
\end{table}
We evaluated our framework on seven models: Qwen2.5\hyp{}1.5B\hyp{}Instruct, Qwen2.5\hyp{}7B\hyp{}Instruct, Granite\hyp{}3.3\hyp{}2B\hyp{}Instruct, Granite\hyp{}3.3\hyp{}8B\hyp{}Instruct, Llama\hyp{}3.2\hyp{}1B\hyp{}Instruct, Llama\hyp{}3.1\hyp{}8B\hyp{}Instruct, and OmniSQL\hyp{}7B. Benchmarks were conducted on the BIRD dev (1,534 samples), Spider dev (1,034 samples), and Spider test (2,147 samples) splits. For both the BIRD and Spider datasets, we began with the original training sets of 9,428 queries (BIRD) and 9,000 queries (Spider). After light pre-processing to remove duplicates and reduce redundancy, we obtained 9,411 queries for BIRD and 8,960 for Spider. For each question, OmniSQL-7B was used to generate $N=32$ candidate SQL queries. We selected $N=32$ following \cite{DBLP:conf/acl/LiL025}, as it represents a well-balanced trade-off between performance and computational cost~\cite{DBLP:conf/iclr/0002WSLCNCZ23,DBLP:conf/nips/LewkowyczADDMRS22}. The framework automatically discards syntactically invalid queries and labels each remaining candidate as correct ($s_+$) or incorrect ($s_-$) depending on whether its result set matches the gold SQL. This procedure yielded final labeled datasets of 82,640 samples for BIRD and 50,073 for Spider.

Analyzing these datasets reveals a marked class imbalance. In Spider, 69.4\% of the candidates are labeled correct, while only 30.6\% are incorrect. BIRD shows a milder skew, with 58.63\% correct and 41.37\% incorrect. To investigate the impact of imbalance on ORM training, we constructed balanced versions of both datasets by downsampling. Specifically, for each question, we retained $\min(s_+, s_-)$ candidates, ensuring a 50/50 split between positive and negative labels. The resulting statistics are summarized in Table~\ref{tab:dataset_stats}.  

With these datasets prepared, we proceed to train ORMs on the seven selected models to address three main research questions:  
\begin{enumerate}[label=(RQ\arabic*), itemsep=1pt, parsep=1pt, topsep=1pt, partopsep=1pt, leftmargin=30pt]
    \item Are ORMs consistently trainable across heterogeneous model families?
    \item How does dataset balancing influence ORM performance?
    \item How effective is the best ORM compared to Majority Voting and execution-based BoN?
    \newline
\end{enumerate}

\begin{table}[t]
\centering
\caption{ORM performance on \textit{BIRD} and \textit{Spider} with the unbalanced dataset, reporting \textit{Execution Accuracy} (\%) on dev and test sets.}
\label{tab:orm_unbalanced}

\footnotesize

\resizebox{\columnwidth}{!}{%
    \begin{tabular}{lccc}
    \toprule
    \textbf{ORM Model} & \textbf{Bird dev} & \textbf{Spider dev} & \textbf{Spider test} \\
    \midrule
    Granite-3.3-2B-Instruct & 66.88 & 82.79 & 85.14 \\
    Granite-3.3-8B-Instruct & 68.25 & 84.42 & 86.91 \\
    \cmidrule(lr){1-4}
    
    Llama-3.2-1B-Instruct   & 67.67 & 84.04 & 86.31 \\
    Llama-3.1-8B-Instruct   & 68.32 & 82.30 & 85.89 \\
    \cmidrule(lr){1-4}
    
    Qwen2.5-1.5B-Instruct   & 68.12 & 84.33 & 86.17 \\
    Qwen2.5-7B-Instruct     & 68.58 & 84.33 & 86.91 \\
    \cmidrule(lr){1-4}
    
    OmniSQL-7B              & 68.64 & 84.42 & 86.63 \\
    \bottomrule
    \end{tabular}
}
\end{table}


\noindent \textbf{RQ1 – Cross-Family Trainability of ORMs.} Table~\ref{tab:orm_unbalanced} reports ORM performance on \textit{BIRD} and \textit{Spider} using the \emph{unbalanced} training set, across Granite, Llama, Qwen, and OmniSQL backbones. The trained ORMs exhibit tightly clustered scores: \textit{Bird dev} ranges from 66.88 to 68.64, \textit{Spider dev} from 82.30 to 84.42, and \textit{Spider test} from 85.14 to 86.91. These close score range shows ORMs train reliably across LLM families.

A closer look reveals three key patterns. The first is that the scale effect is modest: larger variants such as Granite-3.3-8B or Qwen2.5-7B yield only small improvements (typically below two points) over their smaller counterparts, suggesting that ORM effectiveness is not strongly tied to parameter count. The second is family-agnostic generalization: each family contributes at least one competitive ORM, for instance, Granite-3.3-8B and Qwen2.5-7B rank at the top of the \textit{Spider test}, while Llama and OmniSQL models perform comparably on both development and test splits. This indicates that architectural heterogeneity does not hinder successful ORM training. The third is the stability of rankings across splits and datasets: although \textit{Spider} generally achieves higher EX Accuracy than \textit{BIRD}, the relative orderings remain consistent, supporting the robustness of ORMs across evaluation settings.  

Overall, these findings demonstrate that ORMs are \emph{consistently trainable} across heterogeneous LLM families and scales, achieving comparable accuracies with minimal variance. This establishes their generality and motivates the use of any backbone family in the subsequent experiments, RQ2, which investigates the effect of dataset balancing, and RQ3, which benchmarks ORM performance against \textit{test-time} baselines.

\begin{table}[t]
\centering
\caption{Execution accuracy (\%) of ORM models on \textit{BIRD dev}, \textit{Spider dev}, and \textit{Spider test}. Each cell shows results on \emph{Unbalanced} and \emph{Balanced} sets with $\Delta$ (green $\uparrow$ = gain, red $\downarrow$ = drop). Ordered by BIRD Unbalanced; best in \textbf{bold}, runner-up \underline{underlined}.}
\label{tab:orm_comparison_v2_v1_compact}

\resizebox{\columnwidth}{!}{%
    \begin{tabular}{lccc}
    \toprule
    \textbf{ORM Model} &
    \makecell{\textbf{BIRD dev}\\{\scriptsize Unbalanced $\rightarrow$ Balanced}} &
    \makecell{\textbf{Spider dev}\\{\scriptsize Unbalanced $\rightarrow$ Balanced}} &
    \makecell{\textbf{Spider test}\\{\scriptsize Unbalanced $\rightarrow$ Balanced}} \\
    \midrule
    Granite-3.3-2B-Instruct &
      \makecell{66.88 $\rightarrow$ 66.17 \\[-2pt] {\scriptsize\textcolor{BrickRed}{(-0.71 ↓)}}} &
      \makecell{82.79 $\rightarrow$ 82.79 \\[-2pt] {\scriptsize(0.00)}} &
      \makecell{85.14 $\rightarrow$ 85.14 \\[-2pt] {\scriptsize(0.00)}} \\
    \cmidrule(lr){2-4}
    Llama-3.2-1B-Instruct &
      \makecell{67.67 $\rightarrow$ 67.21 \\[-2pt] {\scriptsize\textcolor{BrickRed}{(-0.46 ↓)}}} &
      \makecell{84.04 $\rightarrow$ 83.56 \\[-2pt] {\scriptsize\textcolor{BrickRed}{(-0.48 ↓)}}} &
      \makecell{86.31 $\rightarrow$ 86.54 \\[-2pt] {\scriptsize\textcolor{ForestGreen}{(+0.23 ↑)}}} \\
    \cmidrule(lr){2-4}
    Qwen2.5-1.5B-Instruct &
      \makecell{68.12 $\rightarrow$ 67.28 \\[-2pt] {\scriptsize\textcolor{BrickRed}{(-0.84 ↓)}}} &
      \makecell{84.33 $\rightarrow$ 83.56 \\[-2pt] {\scriptsize\textcolor{BrickRed}{(-0.77 ↓)}}} &
      \makecell{86.17 $\rightarrow$ 86.31 \\[-2pt] {\scriptsize\textcolor{ForestGreen}{(+0.14 ↑)}}} \\
    \cmidrule(lr){2-4}
    Granite-3.3-8B-Instruct &
      \makecell{68.25 $\rightarrow$ 68.38 \\[-2pt] {\scriptsize\textcolor{ForestGreen}{(+0.13 ↑)}}} &
      \makecell{84.42 $\rightarrow$ 83.85 \\[-2pt] {\scriptsize\textcolor{BrickRed}{(-0.57 ↓)}}} &
      \makecell{86.91 $\rightarrow$ \underline{87.00} \\[-2pt] {\scriptsize\textcolor{ForestGreen}{(+0.09 ↑)}}} \\
    \cmidrule(lr){2-4}
    Llama-3.1-8B-Instruct &
      \makecell{68.32 $\rightarrow$ 67.86 \\[-2pt] {\scriptsize\textcolor{BrickRed}{(-0.46 ↓)}}} &
      \makecell{82.30 $\rightarrow$ 84.11 \\[-2pt] {\scriptsize\textcolor{ForestGreen}{(+1.81 ↑)}}} &
      \makecell{85.89 $\rightarrow$ 86.96 \\[-2pt] {\scriptsize\textcolor{ForestGreen}{(+1.07 ↑)}}} \\
    \cmidrule(lr){2-4}
    Qwen2.5-7B-Instruct &
      \makecell{68.58 $\rightarrow$ 68.19 \\[-2pt] {\scriptsize\textcolor{BrickRed}{(-0.39 ↓)}}} &
      \makecell{84.33 $\rightarrow$ 84.14 \\[-2pt] {\scriptsize\textcolor{BrickRed}{(-0.19 ↓)}}} &
      \makecell{86.91 $\rightarrow$ 86.68 \\[-2pt] {\scriptsize\textcolor{BrickRed}{(-0.23 ↓)}}} \\
    \cmidrule(lr){2-4}
    OmniSQL-7B &
      \makecell{\underline{68.64} $\rightarrow$ \textbf{68.90} \\[-2pt] {\scriptsize\textcolor{ForestGreen}{(+0.26 ↑)}}} &
      \makecell{\underline{84.42} $\rightarrow$ \textbf{84.53} \\[-2pt] {\scriptsize\textcolor{ForestGreen}{(+0.11 ↑)}}} &
      \makecell{86.63 $\rightarrow$ \textbf{87.47} \\[-2pt] {\scriptsize\textcolor{ForestGreen}{(+0.84 ↑)}}} \\
    \bottomrule
    \end{tabular}
}
\vspace{-12pt}
\end{table}

\noindent \textbf{RQ2 – The Role of Dataset Balancing.} Table~\ref{tab:orm_comparison_v2_v1_compact} compares ORM performance on \textit{BIRD dev}, \textit{Spider dev}, and \textit{Spider test} when trained on the \emph{Unbalanced} versus \emph{Balanced} datasets. Each cell reports accuracy before and after balancing, with the difference $\Delta$ highlighted in green (↑) for improvements and red (↓) for drops. 

The results reveal a nuanced picture: balancing leads to both gains and losses depending on the backbone and dataset. On \textit{BIRD dev}, several backbones exhibit small decreases, e.g. Qwen2.5-1.5B at $-0.84$ and Qwen2.5-7B at $-0.39$, while Granite-3.3-8B ($+0.13$) and OmniSQL-7B ($+0.26$) show slight improvements. On \textit{Spider dev}, the effect is more mixed: Llama-3.1-8B improves substantially ($+1.81$), whereas others decline moderately, such as Qwen2.5-1.5B ($-0.77$), or remain unchanged, as with Granite-3.3-2B ($0.00$). By contrast, \textit{Spider test} shows more consistent gains, with the largest improvements for Llama-3.1-8B ($+1.07$), OmniSQL-7B ($+0.84$), and Llama-3.2-1B ($+0.23$).

Overall, on \textit{Spider} balancing appears to enhance \emph{generalization}. While accuracy on the development set fluctuates, on the test set balancing lead an overall improvements for the majority of the models. The effect is particularly notable for Llama-3.1-8B and OmniSQL-7B, where gains exceed $+1.0$ and $+0.8$ points. At the same time, the overall relatively small magnitude of changes (generally within $\pm1.2$) suggests that ORMs remain stable even under skewed distributions of labels.

Furthermore, as shown in Table~\ref{tab:dataset_stats}, the number of training samples differs substantially: the imbalanced sets contain over 80k (\textit{BIRD train}) and 50k (\textit{Spider train}) examples, while the balanced counterparts reduce to just 30k and 18k, respectively. This reduction makes training considerably faster, lowering both computational cost and time-to-train. Despite the smaller datasets, ORM performance remains largely comparable across settings, with overall differences being minimal. More interestingly, in certain cases balancing even improves results, for instance, the best-performing models on \textit{BIRD dev}, such as OmniSQL-7B, achieve higher execution accuracy when trained on the smaller, balanced set. 

In summary, dataset balancing not only reduces data requirements and training time, but also yields competitive, and sometimes superior, performance. Based on these trends, we adopt the \emph{Balanced} configuration as the default setup for the comparison against baselines in RQ3.

\noindent \textbf{RQ3 – ORM Effectiveness vs. Baseline Methods.} 

\begin{table}[t]
\centering
\caption{Execution accuracy (\%) of OmniSQL-7B on \textit{BIRD dev}, \textit{Spider dev}, and \textit{Spider test}. 
$\Delta$ indicates the gain over the baseline (\emph{$N{=}1$}). 
Best and runner-up results are in \textbf{bold} and \underline{underlined}, respectively. 
Values marked with $\dagger$ are not McNemar significant ($p<0.05$), and those with * lose significance after Bonferroni correction.}

\label{tab:main_results}



\resizebox{\columnwidth}{!}{%
  \setlength{\tabcolsep}{5pt}
  \begin{tabular}{lcccccc}
    \toprule
    \multirow{2}{*}{\textbf{Method}} &
      \multicolumn{2}{c}{\textbf{BIRD dev}} &
      \multicolumn{2}{c}{\textbf{Spider dev}} &
      \multicolumn{2}{c}{\textbf{Spider test}} \\
    \cmidrule(lr){2-3}\cmidrule(lr){4-5}\cmidrule(lr){6-7}
    & EX & $\Delta$ & EX & $\Delta$ & EX & $\Delta$ \\
    \midrule
    Baseline (N=1) & 63.89 & -- & 82.40 & -- & 84.02 & -- \\
    Majority Voting (N=32)  & \underline{66.95} & +3.06 & \underline{83.75}$*$ & +1.35 & \underline{85.47} & +1.45 \\
    \multicolumn{7}{l}{Best-of-$N$ (N=32)} \\
    \quad Execution-based & 66.04 & +2.15 & 82.79$\dagger*$ & +0.39 & 85.14 & +1.12 \\
    \quad ORM-based & \textbf{68.90} & +5.01 & \textbf{84.53}$*$ & +2.13 & \textbf{87.47} & +3.45 \\
    \bottomrule
  \end{tabular}%
}
\end{table}

Table~\ref{tab:main_results} compares the ORM-based Best-of-$N$ strategy with execution\hyp{}based BoN (ex-BoN) and Majority Voting (Maj), using OmniSQL-7B as the candidate generator. The ORM is trained on OmniSQL-7B with the balanced dataset. We report execution accuracy on \textit{BIRD dev}, \textit{Spider dev}, and \textit{Spider test}, along with the absolute gains over the single-sample baseline ($N=1$), corresponding to the first generated query. Following~\cite{DBLP:conf/acl/ReichartDBS18}, we apply the McNemar statistical test together with the Bonferroni correction to assess significance across methods. Results indicate that the ORM-based strategy consistently achieves the highest execution accuracy across all datasets and evaluation splits. Although the improvement on \textit{Spider dev} does not retain significance after Bonferroni correction, which may be attributed to the smaller sample size, all other results remain statistically significant

Specifically, on \textit{BIRD dev}, ORMs reach 68.90\% execution accuracy, surpassing both Maj (66.95\%) and ex-BoN (66.04\%), and achieving a gain of +5.01 points over the baseline. On \textit{Spider dev}, ORMs again outperform all baselines with 84.53\% accuracy, a +2.13 improvement compared to 83.75\% for Maj and 82.79\% for ex-BoN. The same trend holds on \textit{Spider test}, where ORM-based selection achieves 87.47\%, outperforming both Maj (85.47\%) and ex-BoN (85.14\%), with an absolute gain of +3.45 points over the baseline.

These results highlight a clear pattern: while heuristic strategies such as Majority Voting and execution-based Best-of-$N$ offer gains over the single-sample baseline, ORM-based selection delivers consistent and stronger improvements. Notably, the largest gain is observed on \textit{BIRD dev} (+5.01), indicating that ORMs provide the greatest advantage in scenarios with more challenging or diverse query distributions. 



Overall, these findings demonstrate that ORMs not only generalize across model families (RQ1) and benefit from dataset balancing (RQ2), but also achieve improvements over widely used test-time strategies. This represents the central contribution of our work: showing that ORM-based selection outperforms Majority Voting and execution-based Best-of-$N$, providing a stronger heuristic for Best-of-$N$ test-time inference.

\section{Ablation Studies}
In this section, we further investigate ORM training by addressing additional research questions to better understand its effectiveness compared to alternative methods. Specifically, we aim to answer:  
\begin{enumerate}[label=(RQ\arabic*), start=4, itemsep=1pt, parsep=1pt, topsep=1pt, partopsep=1pt, leftmargin=30pt]
    \item What is the role of $N$ in ORM performance?
    \item Are ORMs effective when using larger base models as generators?
    \item How does the training prompt influence ORM results?
    \item Are ORM sizes consistent with previous findings when scaling beyond 7B models?
    \item Is autoregressive fine-tuning the best choice for training an ORM?
\end{enumerate}

\begin{table*}[t]
\centering
\caption{Execution accuracy (\%) and Pass@$N$ of test-time strategies (execution-based BoN, Majority Voting, ORM-based BoN) across varying $N$ values on the \textit{BIRD dev}, \textit{Spider dev}, and \textit{Spider test} benchmarks. Columns are color-coded from light yellow (lowest) to dark orange (highest) performance.}
\label{tab:n_decreasing}
\renewcommand{\arraystretch}{1.05}
\resizebox{0.9\textwidth}{!}{%
    \begin{tabular}{c
                    ccc
                    ccc
                    ccc
                    ccc}
    \toprule
    \multirow{2}{*}{$N$} 
    & \multicolumn{3}{c}{\textbf{Execution-based BoN}}
    & \multicolumn{3}{c}{\textbf{Majority Voting}} 
    & \multicolumn{3}{c}{\textbf{ORM-based BoN}} 
    & \multicolumn{3}{c}{\textbf{Pass@$N$}} \\ 
    \cmidrule(lr){2-4}\cmidrule(lr){5-7}\cmidrule(lr){8-10}\cmidrule(lr){11-13}
    & BIRD dev & Spider dev & Spider test 
    & BIRD dev & Spider dev & Spider test 
    & BIRD dev & Spider dev & Spider test 
    & BIRD dev & Spider dev & Spider test \\
    \midrule
    32 & \cellcolor{heatmax!100!heatmin} 66.04 & \cellcolor{heatmax!81!heatmin} 82.79 & \cellcolor{heatmax!100!heatmin} 85.14 & \cellcolor{heatmax!94!heatmin} 66.95 & \cellcolor{heatmax!78!heatmin} 83.75 & \cellcolor{heatmax!84!heatmin} 85.47 & \cellcolor{heatmax!100!heatmin} 68.90 & \cellcolor{heatmax!92!heatmin} 84.53 & \cellcolor{heatmax!100!heatmin} 87.47 & \cellcolor{heatmax!100!heatmin} 80.57 & \cellcolor{heatmax!100!heatmin} 91.68 & \cellcolor{heatmax!100!heatmin} 93.29 \\
    31 & \cellcolor{heatmax!100!heatmin} 66.04 & \cellcolor{heatmax!81!heatmin} 82.79 & \cellcolor{heatmax!100!heatmin} 85.14 & \cellcolor{heatmax!98!heatmin} 67.08 & \cellcolor{heatmax!78!heatmin} 83.75 & \cellcolor{heatmax!84!heatmin} 85.47 & \cellcolor{heatmax!94!heatmin} 68.58 & \cellcolor{heatmax!92!heatmin} 84.53 & \cellcolor{heatmax!100!heatmin} 87.47 & \cellcolor{heatmax!98!heatmin} 80.25 & \cellcolor{heatmax!100!heatmin} 91.68 & \cellcolor{heatmax!99!heatmin} 93.20 \\
    30 & \cellcolor{heatmax!100!heatmin} 66.04 & \cellcolor{heatmax!100!heatmin} 82.88 & \cellcolor{heatmax!100!heatmin} 85.14 & \cellcolor{heatmax!96!heatmin} 67.01 & \cellcolor{heatmax!72!heatmin} 83.66 & \cellcolor{heatmax!81!heatmin} 85.42 & \cellcolor{heatmax!94!heatmin} 68.58 & \cellcolor{heatmax!96!heatmin} 84.62 & \cellcolor{heatmax!100!heatmin} 87.47 & \cellcolor{heatmax!98!heatmin} 80.25 & \cellcolor{heatmax!100!heatmin} 91.68 & \cellcolor{heatmax!98!heatmin} 93.15 \\
    29 & \cellcolor{heatmax!97!heatmin} 65.97 & \cellcolor{heatmax!100!heatmin} 82.88 & \cellcolor{heatmax!100!heatmin} 85.14 & \cellcolor{heatmax!94!heatmin} 66.95 & \cellcolor{heatmax!67!heatmin} 83.56 & \cellcolor{heatmax!81!heatmin} 85.42 & \cellcolor{heatmax!92!heatmin} 68.51 & \cellcolor{heatmax!96!heatmin} 84.62 & \cellcolor{heatmax!99!heatmin} 87.42 & \cellcolor{heatmax!97!heatmin} 80.12 & \cellcolor{heatmax!100!heatmin} 91.68 & \cellcolor{heatmax!98!heatmin} 93.11 \\
    28 & \cellcolor{heatmax!97!heatmin} 65.97 & \cellcolor{heatmax!100!heatmin} 82.88 & \cellcolor{heatmax!100!heatmin} 85.14 & \cellcolor{heatmax!96!heatmin} 67.01 & \cellcolor{heatmax!78!heatmin} 83.75 & \cellcolor{heatmax!84!heatmin} 85.47 & \cellcolor{heatmax!94!heatmin} 68.58 & \cellcolor{heatmax!96!heatmin} 84.62 & \cellcolor{heatmax!97!heatmin} 87.38 & \cellcolor{heatmax!97!heatmin} 80.05 & \cellcolor{heatmax!100!heatmin} 91.68 & \cellcolor{heatmax!97!heatmin} 93.01 \\
    27 & \cellcolor{heatmax!97!heatmin} 65.97 & \cellcolor{heatmax!100!heatmin} 82.88 & \cellcolor{heatmax!100!heatmin} 85.14 & \cellcolor{heatmax!100!heatmin} 67.14 & \cellcolor{heatmax!83!heatmin} 83.85 & \cellcolor{heatmax!84!heatmin} 85.47 & \cellcolor{heatmax!94!heatmin} 68.58 & \cellcolor{heatmax!96!heatmin} 84.62 & \cellcolor{heatmax!97!heatmin} 87.38 & \cellcolor{heatmax!96!heatmin} 79.92 & \cellcolor{heatmax!99!heatmin} 91.59 & \cellcolor{heatmax!97!heatmin} 93.01 \\
    26 & \cellcolor{heatmax!97!heatmin} 65.97 & \cellcolor{heatmax!100!heatmin} 82.88 & \cellcolor{heatmax!100!heatmin} 85.14 & \cellcolor{heatmax!88!heatmin} 66.75 & \cellcolor{heatmax!78!heatmin} 83.75 & \cellcolor{heatmax!76!heatmin} 85.33 & \cellcolor{heatmax!95!heatmin} 68.64 & \cellcolor{heatmax!96!heatmin} 84.62 & \cellcolor{heatmax!97!heatmin} 87.38 & \cellcolor{heatmax!96!heatmin} 79.92 & \cellcolor{heatmax!99!heatmin} 91.59 & \cellcolor{heatmax!95!heatmin} 92.87 \\
    25 & \cellcolor{heatmax!97!heatmin} 65.97 & \cellcolor{heatmax!100!heatmin} 82.88 & \cellcolor{heatmax!100!heatmin} 85.14 & \cellcolor{heatmax!88!heatmin} 66.75 & \cellcolor{heatmax!83!heatmin} 83.85 & \cellcolor{heatmax!86!heatmin} 85.51 & \cellcolor{heatmax!96!heatmin} 68.71 & \cellcolor{heatmax!96!heatmin} 84.62 & \cellcolor{heatmax!97!heatmin} 87.38 & \cellcolor{heatmax!95!heatmin} 79.79 & \cellcolor{heatmax!99!heatmin} 91.59 & \cellcolor{heatmax!95!heatmin} 92.87 \\
    24 & \cellcolor{heatmax!97!heatmin} 65.97 & \cellcolor{heatmax!100!heatmin} 82.88 & \cellcolor{heatmax!100!heatmin} 85.14 & \cellcolor{heatmax!96!heatmin} 67.01 & \cellcolor{heatmax!89!heatmin} 83.95 & \cellcolor{heatmax!89!heatmin} 85.56 & \cellcolor{heatmax!97!heatmin} 68.77 & \cellcolor{heatmax!96!heatmin} 84.62 & \cellcolor{heatmax!96!heatmin} 87.33 & \cellcolor{heatmax!94!heatmin} 79.53 & \cellcolor{heatmax!99!heatmin} 91.59 & \cellcolor{heatmax!94!heatmin} 92.78 \\
    23 & \cellcolor{heatmax!97!heatmin} 65.97 & \cellcolor{heatmax!100!heatmin} 82.88 & \cellcolor{heatmax!100!heatmin} 85.14 & \cellcolor{heatmax!98!heatmin} 67.08 & \cellcolor{heatmax!100!heatmin} 84.14 & \cellcolor{heatmax!84!heatmin} 85.47 & \cellcolor{heatmax!96!heatmin} 68.71 & \cellcolor{heatmax!96!heatmin} 84.62 & \cellcolor{heatmax!96!heatmin} 87.33 & \cellcolor{heatmax!93!heatmin} 79.47 & \cellcolor{heatmax!98!heatmin} 91.49 & \cellcolor{heatmax!94!heatmin} 92.78 \\
    22 & \cellcolor{heatmax!97!heatmin} 65.97 & \cellcolor{heatmax!100!heatmin} 82.88 & \cellcolor{heatmax!100!heatmin} 85.14 & \cellcolor{heatmax!92!heatmin} 66.88 & \cellcolor{heatmax!89!heatmin} 83.95 & \cellcolor{heatmax!81!heatmin} 85.42 & \cellcolor{heatmax!100!heatmin} 68.90 & \cellcolor{heatmax!96!heatmin} 84.62 & \cellcolor{heatmax!96!heatmin} 87.33 & \cellcolor{heatmax!93!heatmin} 79.33 & \cellcolor{heatmax!97!heatmin} 91.39 & \cellcolor{heatmax!94!heatmin} 92.73 \\
    21 & \cellcolor{heatmax!94!heatmin} 65.91 & \cellcolor{heatmax!100!heatmin} 82.88 & \cellcolor{heatmax!100!heatmin} 85.14 & \cellcolor{heatmax!88!heatmin} 66.75 & \cellcolor{heatmax!83!heatmin} 83.85 & \cellcolor{heatmax!84!heatmin} 85.47 & \cellcolor{heatmax!99!heatmin} 68.84 & \cellcolor{heatmax!96!heatmin} 84.62 & \cellcolor{heatmax!94!heatmin} 87.28 & \cellcolor{heatmax!91!heatmin} 79.14 & \cellcolor{heatmax!95!heatmin} 91.20 & \cellcolor{heatmax!94!heatmin} 92.73 \\
    20 & \cellcolor{heatmax!94!heatmin} 65.91 & \cellcolor{heatmax!100!heatmin} 82.88 & \cellcolor{heatmax!100!heatmin} 85.14 & \cellcolor{heatmax!96!heatmin} 67.01 & \cellcolor{heatmax!78!heatmin} 83.75 & \cellcolor{heatmax!89!heatmin} 85.56 & \cellcolor{heatmax!97!heatmin} 68.77 & \cellcolor{heatmax!96!heatmin} 84.62 & \cellcolor{heatmax!96!heatmin} 87.33 & \cellcolor{heatmax!91!heatmin} 79.07 & \cellcolor{heatmax!93!heatmin} 91.01 & \cellcolor{heatmax!94!heatmin} 92.69 \\
    19 & \cellcolor{heatmax!94!heatmin} 65.91 & \cellcolor{heatmax!100!heatmin} 82.88 & \cellcolor{heatmax!100!heatmin} 85.14 & \cellcolor{heatmax!98!heatmin} 67.08 & \cellcolor{heatmax!83!heatmin} 83.85 & \cellcolor{heatmax!78!heatmin} 85.37 & \cellcolor{heatmax!97!heatmin} 68.77 & \cellcolor{heatmax!96!heatmin} 84.62 & \cellcolor{heatmax!96!heatmin} 87.33 & \cellcolor{heatmax!90!heatmin} 78.88 & \cellcolor{heatmax!92!heatmin} 90.91 & \cellcolor{heatmax!92!heatmin} 92.59 \\
    18 & \cellcolor{heatmax!94!heatmin} 65.91 & \cellcolor{heatmax!100!heatmin} 82.88 & \cellcolor{heatmax!100!heatmin} 85.14 & \cellcolor{heatmax!96!heatmin} 67.01 & \cellcolor{heatmax!78!heatmin} 83.76 & \cellcolor{heatmax!94!heatmin} 85.65 & \cellcolor{heatmax!99!heatmin} 68.84 & \cellcolor{heatmax!96!heatmin} 84.62 & \cellcolor{heatmax!97!heatmin} 87.38 & \cellcolor{heatmax!90!heatmin} 78.88 & \cellcolor{heatmax!91!heatmin} 90.81 & \cellcolor{heatmax!92!heatmin} 92.55 \\
    17 & \cellcolor{heatmax!94!heatmin} 65.91 & \cellcolor{heatmax!100!heatmin} 82.88 & \cellcolor{heatmax!100!heatmin} 85.14 & \cellcolor{heatmax!98!heatmin} 67.08 & \cellcolor{heatmax!89!heatmin} 83.95 & \cellcolor{heatmax!92!heatmin} 85.61 & \cellcolor{heatmax!97!heatmin} 68.77 & \cellcolor{heatmax!100!heatmin} 84.72 & \cellcolor{heatmax!96!heatmin} 87.33 & \cellcolor{heatmax!89!heatmin} 78.68 & \cellcolor{heatmax!91!heatmin} 90.81 & \cellcolor{heatmax!91!heatmin} 92.50 \\
    16 & \cellcolor{heatmax!94!heatmin} 65.91 & \cellcolor{heatmax!100!heatmin} 82.88 & \cellcolor{heatmax!100!heatmin} 85.14 & \cellcolor{heatmax!94!heatmin} 66.95 & \cellcolor{heatmax!72!heatmin} 83.66 & \cellcolor{heatmax!92!heatmin} 85.61 & \cellcolor{heatmax!97!heatmin} 68.77 & \cellcolor{heatmax!92!heatmin} 84.53 & \cellcolor{heatmax!96!heatmin} 87.33 & \cellcolor{heatmax!88!heatmin} 78.49 & \cellcolor{heatmax!90!heatmin} 90.72 & \cellcolor{heatmax!90!heatmin} 92.40 \\
    15 & \cellcolor{heatmax!91!heatmin} 65.84 & \cellcolor{heatmax!100!heatmin} 82.88 & \cellcolor{heatmax!100!heatmin} 85.14 & \cellcolor{heatmax!100!heatmin} 67.14 & \cellcolor{heatmax!83!heatmin} 83.85 & \cellcolor{heatmax!81!heatmin} 85.42 & \cellcolor{heatmax!96!heatmin} 68.71 & \cellcolor{heatmax!92!heatmin} 84.53 & \cellcolor{heatmax!94!heatmin} 87.28 & \cellcolor{heatmax!86!heatmin} 78.23 & \cellcolor{heatmax!87!heatmin} 90.52 & \cellcolor{heatmax!89!heatmin} 92.27 \\
    14 & \cellcolor{heatmax!91!heatmin} 65.84 & \cellcolor{heatmax!100!heatmin} 82.88 & \cellcolor{heatmax!96!heatmin} 85.10 & \cellcolor{heatmax!96!heatmin} 67.01 & \cellcolor{heatmax!83!heatmin} 83.85 & \cellcolor{heatmax!92!heatmin} 85.61 & \cellcolor{heatmax!95!heatmin} 68.64 & \cellcolor{heatmax!92!heatmin} 84.53 & \cellcolor{heatmax!92!heatmin} 87.19 & \cellcolor{heatmax!84!heatmin} 77.90 & \cellcolor{heatmax!87!heatmin} 90.52 & \cellcolor{heatmax!87!heatmin} 92.13 \\
    13 & \cellcolor{heatmax!91!heatmin} 65.84 & \cellcolor{heatmax!100!heatmin} 82.88 & \cellcolor{heatmax!96!heatmin} 85.10 & \cellcolor{heatmax!100!heatmin} 67.14 & \cellcolor{heatmax!72!heatmin} 83.66 & \cellcolor{heatmax!89!heatmin} 85.56 & \cellcolor{heatmax!95!heatmin} 68.64 & \cellcolor{heatmax!92!heatmin} 84.53 & \cellcolor{heatmax!92!heatmin} 87.19 & \cellcolor{heatmax!84!heatmin} 77.84 & \cellcolor{heatmax!87!heatmin} 90.43 & \cellcolor{heatmax!87!heatmin} 92.08 \\
    12 & \cellcolor{heatmax!91!heatmin} 65.84 & \cellcolor{heatmax!100!heatmin} 82.88 & \cellcolor{heatmax!92!heatmin} 85.05 & \cellcolor{heatmax!92!heatmin} 66.88 & \cellcolor{heatmax!83!heatmin} 83.85 & \cellcolor{heatmax!86!heatmin} 85.51 & \cellcolor{heatmax!96!heatmin} 68.71 & \cellcolor{heatmax!88!heatmin} 84.44 & \cellcolor{heatmax!85!heatmin} 86.96 & \cellcolor{heatmax!83!heatmin} 77.71 & \cellcolor{heatmax!87!heatmin} 90.43 & \cellcolor{heatmax!85!heatmin} 91.94 \\
    11 & \cellcolor{heatmax!91!heatmin} 65.84 & \cellcolor{heatmax!100!heatmin} 82.88 & \cellcolor{heatmax!92!heatmin} 85.05 & \cellcolor{heatmax!90!heatmin} 66.82 & \cellcolor{heatmax!72!heatmin} 83.66 & \cellcolor{heatmax!68!heatmin} 85.19 & \cellcolor{heatmax!95!heatmin} 68.64 & \cellcolor{heatmax!83!heatmin} 84.33 & \cellcolor{heatmax!83!heatmin} 86.87 & \cellcolor{heatmax!80!heatmin} 77.25 & \cellcolor{heatmax!84!heatmin} 90.23 & \cellcolor{heatmax!83!heatmin} 91.71 \\
    10 & \cellcolor{heatmax!85!heatmin} 65.71 & \cellcolor{heatmax!100!heatmin} 82.88 & \cellcolor{heatmax!92!heatmin} 85.05 & \cellcolor{heatmax!86!heatmin} 66.69 & \cellcolor{heatmax!100!heatmin} 84.14 & \cellcolor{heatmax!89!heatmin} 85.56 & \cellcolor{heatmax!95!heatmin} 68.64 & \cellcolor{heatmax!83!heatmin} 84.33 & \cellcolor{heatmax!84!heatmin} 86.91 & \cellcolor{heatmax!79!heatmin} 76.99 & \cellcolor{heatmax!82!heatmin} 90.04 & \cellcolor{heatmax!82!heatmin} 91.66 \\
    9 & \cellcolor{heatmax!85!heatmin} 65.71 & \cellcolor{heatmax!100!heatmin} 82.88 & \cellcolor{heatmax!92!heatmin} 85.05 & \cellcolor{heatmax!80!heatmin} 66.49 & \cellcolor{heatmax!94!heatmin} 84.04 & \cellcolor{heatmax!94!heatmin} 85.65 & \cellcolor{heatmax!92!heatmin} 68.51 & \cellcolor{heatmax!79!heatmin} 84.24 & \cellcolor{heatmax!84!heatmin} 86.91 & \cellcolor{heatmax!76!heatmin} 76.60 & \cellcolor{heatmax!80!heatmin} 89.85 & \cellcolor{heatmax!82!heatmin} 91.62 \\
    8 & \cellcolor{heatmax!85!heatmin} 65.71 & \cellcolor{heatmax!100!heatmin} 82.88 & \cellcolor{heatmax!92!heatmin} 85.05 & \cellcolor{heatmax!94!heatmin} 66.95 & \cellcolor{heatmax!89!heatmin} 83.95 & \cellcolor{heatmax!92!heatmin} 85.61 & \cellcolor{heatmax!92!heatmin} 68.51 & \cellcolor{heatmax!83!heatmin} 84.33 & \cellcolor{heatmax!81!heatmin} 86.82 & \cellcolor{heatmax!73!heatmin} 76.14 & \cellcolor{heatmax!78!heatmin} 89.65 & \cellcolor{heatmax!80!heatmin} 91.48 \\
    7 & \cellcolor{heatmax!79!heatmin} 65.58 & \cellcolor{heatmax!81!heatmin} 82.79 & \cellcolor{heatmax!92!heatmin} 85.05 & \cellcolor{heatmax!84!heatmin} 66.62 & \cellcolor{heatmax!94!heatmin} 84.04 & \cellcolor{heatmax!97!heatmin} 85.70 & \cellcolor{heatmax!90!heatmin} 68.38 & \cellcolor{heatmax!83!heatmin} 84.33 & \cellcolor{heatmax!80!heatmin} 86.77 & \cellcolor{heatmax!69!heatmin} 75.36 & \cellcolor{heatmax!74!heatmin} 89.26 & \cellcolor{heatmax!76!heatmin} 91.06 \\
    6 & \cellcolor{heatmax!73!heatmin} 65.45 & \cellcolor{heatmax!81!heatmin} 82.79 & \cellcolor{heatmax!88!heatmin} 85.00 & \cellcolor{heatmax!78!heatmin} 66.43 & \cellcolor{heatmax!72!heatmin} 83.66 & \cellcolor{heatmax!100!heatmin} 85.75 & \cellcolor{heatmax!79!heatmin} 67.86 & \cellcolor{heatmax!79!heatmin} 84.24 & \cellcolor{heatmax!73!heatmin} 86.54 & \cellcolor{heatmax!64!heatmin} 74.58 & \cellcolor{heatmax!73!heatmin} 89.17 & \cellcolor{heatmax!73!heatmin} 90.82 \\
    5 & \cellcolor{heatmax!69!heatmin} 65.38 & \cellcolor{heatmax!81!heatmin} 82.79 & \cellcolor{heatmax!88!heatmin} 85.00 & \cellcolor{heatmax!64!heatmin} 65.97 & \cellcolor{heatmax!89!heatmin} 83.95 & \cellcolor{heatmax!94!heatmin} 85.65 & \cellcolor{heatmax!71!heatmin} 67.47 & \cellcolor{heatmax!75!heatmin} 84.14 & \cellcolor{heatmax!72!heatmin} 86.49 & \cellcolor{heatmax!59!heatmin} 73.79 & \cellcolor{heatmax!67!heatmin} 88.59 & \cellcolor{heatmax!70!heatmin} 90.54 \\
    4 & \cellcolor{heatmax!67!heatmin} 65.32 & \cellcolor{heatmax!100!heatmin} 82.88 & \cellcolor{heatmax!88!heatmin} 85.00 & \cellcolor{heatmax!44!heatmin} 65.32 & \cellcolor{heatmax!72!heatmin} 83.66 & \cellcolor{heatmax!84!heatmin} 85.47 & \cellcolor{heatmax!69!heatmin} 67.34 & \cellcolor{heatmax!71!heatmin} 84.04 & \cellcolor{heatmax!72!heatmin} 86.49 & \cellcolor{heatmax!54!heatmin} 72.82 & \cellcolor{heatmax!64!heatmin} 88.30 & \cellcolor{heatmax!66!heatmin} 90.13 \\
    3 & \cellcolor{heatmax!57!heatmin} 65.12 & \cellcolor{heatmax!100!heatmin} 82.88 & \cellcolor{heatmax!79!heatmin} 84.91 & \cellcolor{heatmax!40!heatmin} 65.19 & \cellcolor{heatmax!67!heatmin} 83.56 & \cellcolor{heatmax!76!heatmin} 85.33 & \cellcolor{heatmax!57!heatmin} 66.75 & \cellcolor{heatmax!71!heatmin} 84.04 & \cellcolor{heatmax!62!heatmin} 86.17 & \cellcolor{heatmax!42!heatmin} 70.86 & \cellcolor{heatmax!54!heatmin} 87.43 & \cellcolor{heatmax!55!heatmin} 89.10 \\
    2 & \cellcolor{heatmax!45!heatmin} 64.86 & \cellcolor{heatmax!81!heatmin} 82.79 & \cellcolor{heatmax!54!heatmin} 84.63 & \cellcolor{heatmax!16!heatmin} 64.41 & \cellcolor{heatmax!39!heatmin} 83.08 & \cellcolor{heatmax!24!heatmin} 84.44 & \cellcolor{heatmax!43!heatmin} 66.04 & \cellcolor{heatmax!54!heatmin} 83.66 & \cellcolor{heatmax!50!heatmin} 85.75 & \cellcolor{heatmax!30!heatmin} 68.84 & \cellcolor{heatmax!40!heatmin} 86.07 & \cellcolor{heatmax!39!heatmin} 87.66 \\
    1 & \cellcolor{heatmax!0!heatmin} 63.89 & \cellcolor{heatmax!0!heatmin} 82.40 & \cellcolor{heatmax!0!heatmin} 84.02 & \cellcolor{heatmax!0!heatmin} 63.89 & \cellcolor{heatmax!0!heatmin} 82.40 & \cellcolor{heatmax!0!heatmin} 84.02 & \cellcolor{heatmax!0!heatmin} 63.89 & \cellcolor{heatmax!0!heatmin} 82.40 & \cellcolor{heatmax!0!heatmin} 84.02 & \cellcolor{heatmax!0!heatmin} 63.89 & \cellcolor{heatmax!0!heatmin} 82.40 & \cellcolor{heatmax!0!heatmin} 84.02 \\
    \bottomrule
    \end{tabular}
}
\end{table*}
\begin{figure*}[t]
  \centering
  \subfloat[Simple EX at scaling $N$.]{%
    \includegraphics[width=0.24\textwidth]{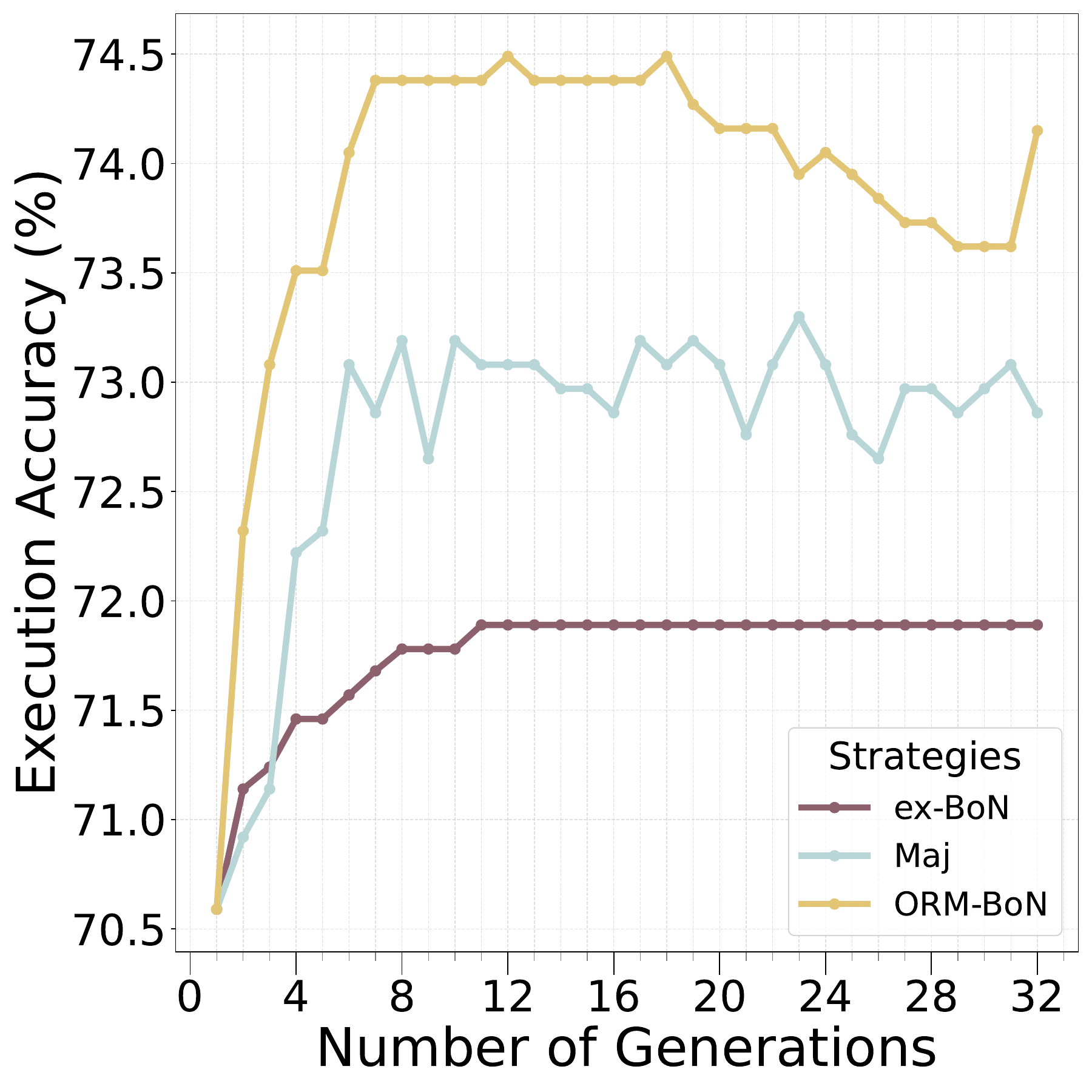}%
    \label{fig:simple_ex}
  }\hfill
  \subfloat[Moderate EX at scaling $N$.]{%
    \includegraphics[width=0.24\textwidth]{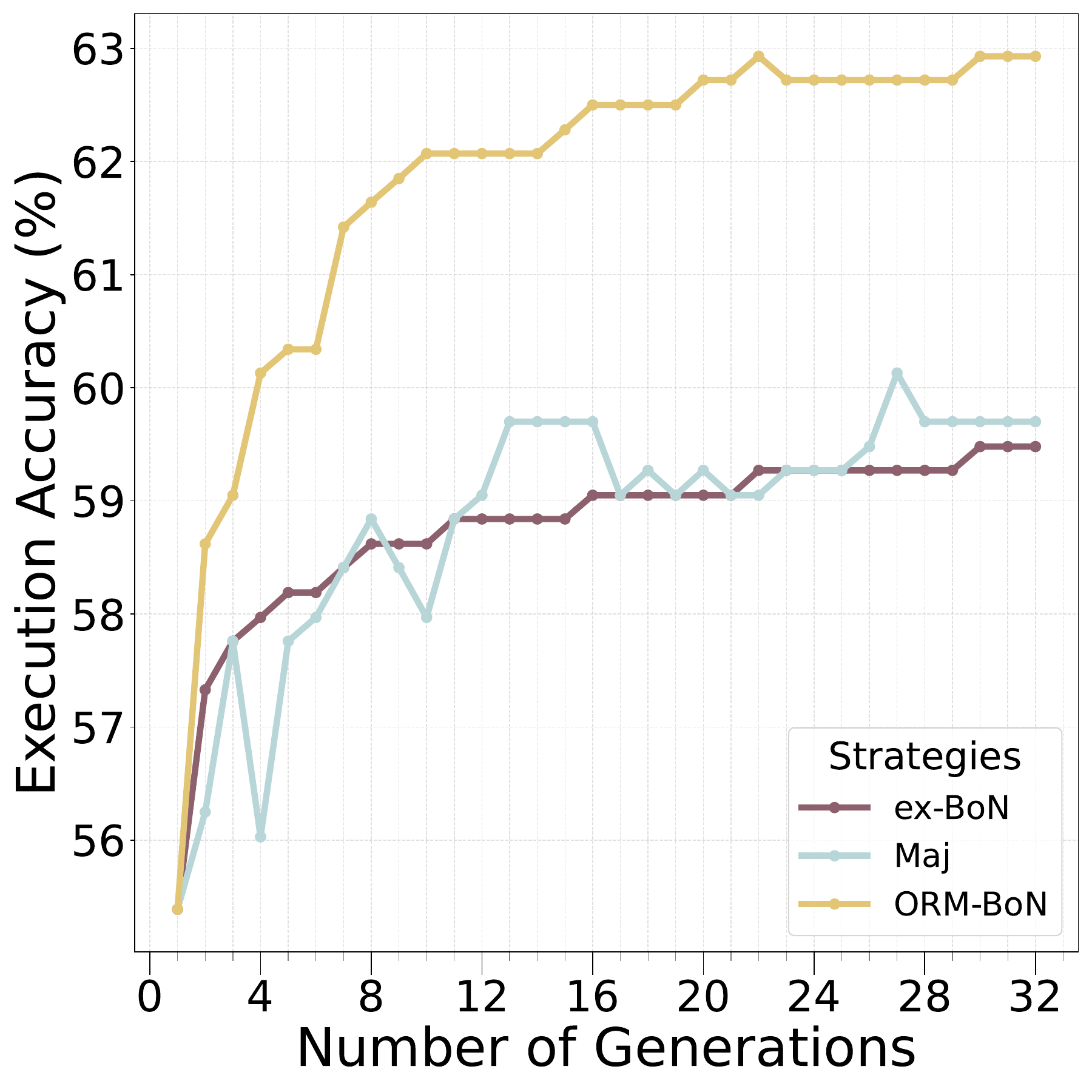}%
    \label{fig:moderate_ex}
  }\hfill
  \subfloat[Challenging EX at scaling $N$.]{%
    \includegraphics[width=0.24\textwidth]{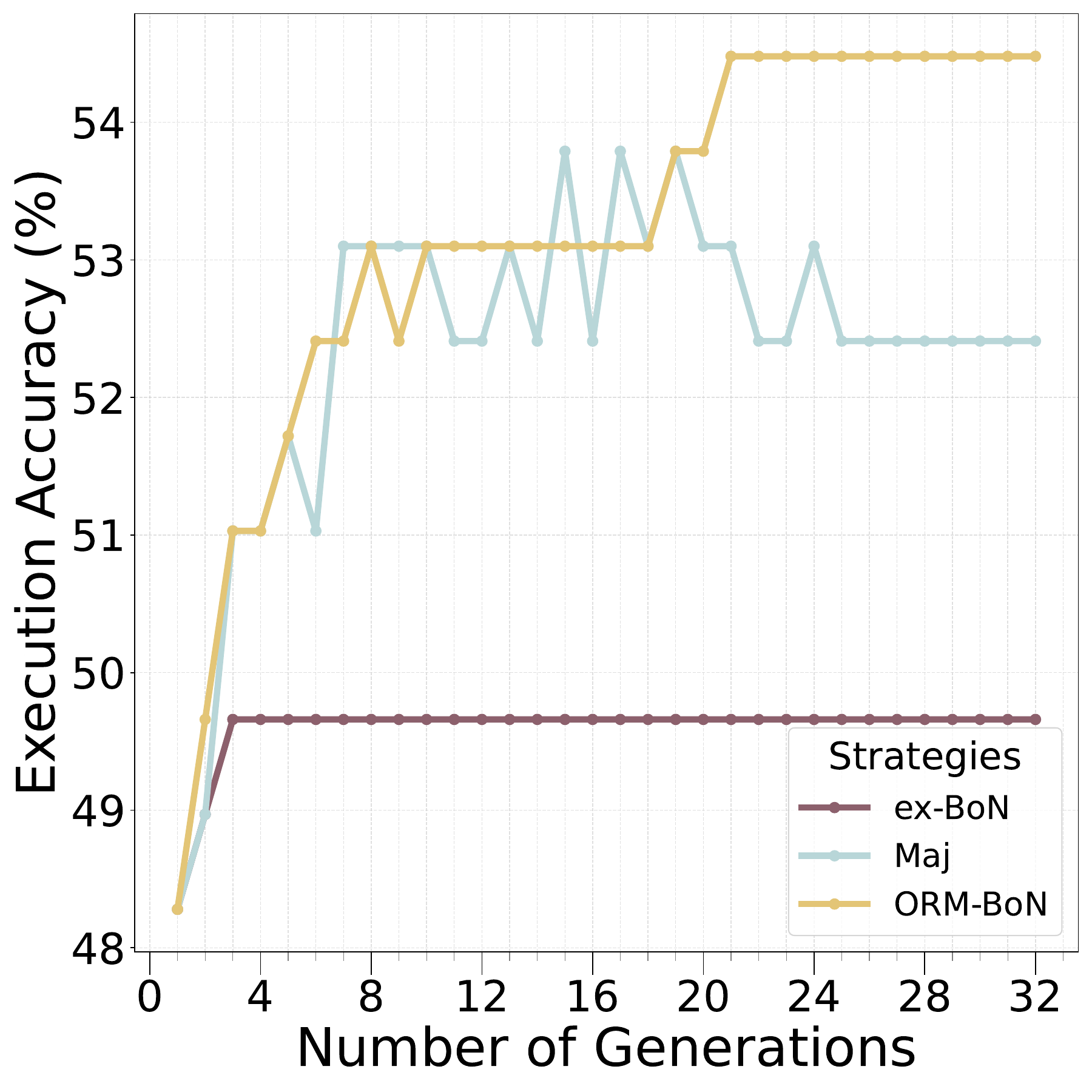}%
    \label{fig:challenging_ex}
  }\hfill
  \subfloat[Total EX at scaling $N$.]{%
    \includegraphics[width=0.24\textwidth]{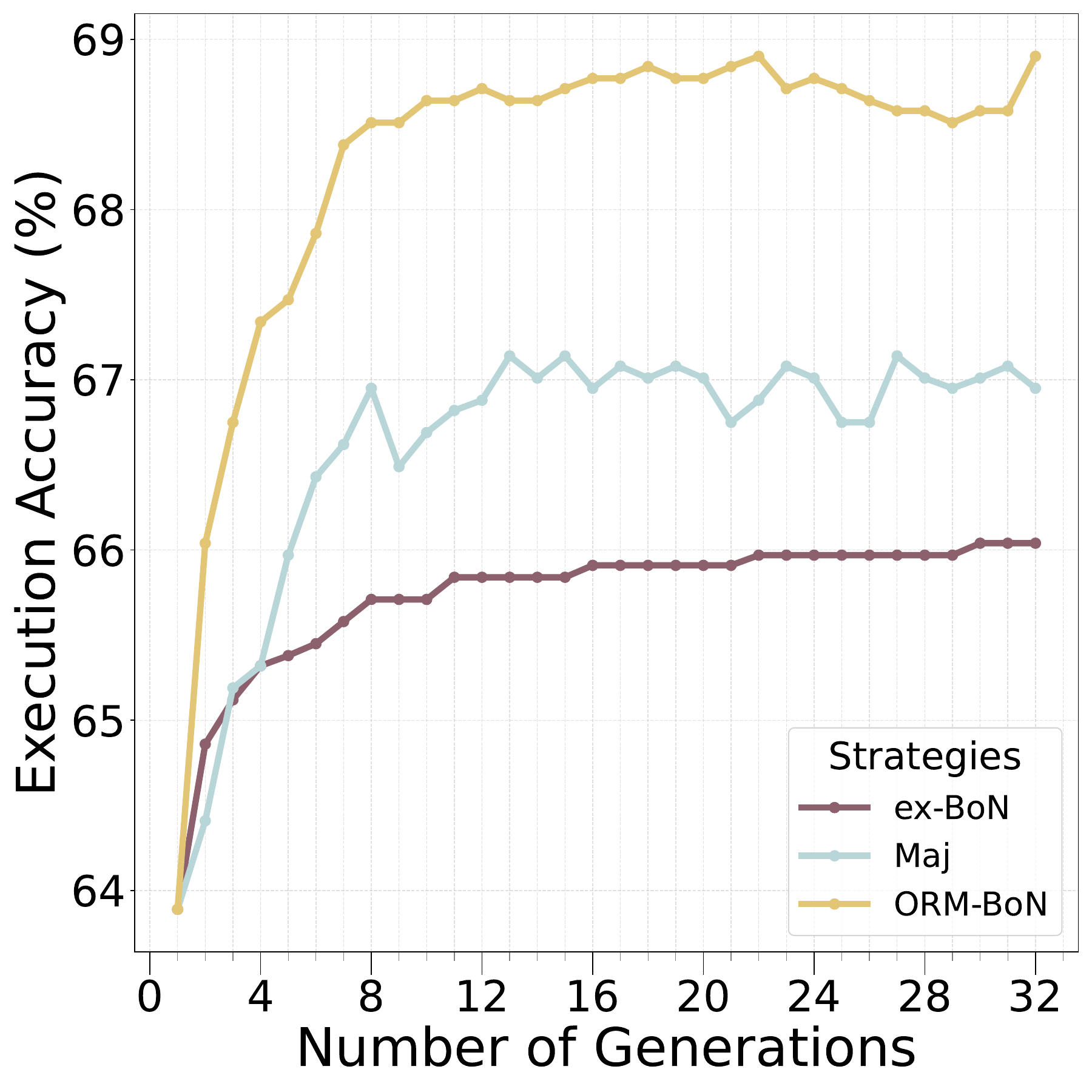}%
    \label{fig:total_ex}
  }

  \caption{Execution accuracy (\%) on \textit{BIRD dev} across different $N$ values, comparing execution-based Best-of-$N$ (ex-BoN), Majority Voting (Maj), and ORM-based Best-of-$N$ (ORM-BoN). Results are shown by query difficulty (Simple, Moderate, Challenging) and overall (Total)}
  \label{fig:n_decreasing_difficulty}
\end{figure*}

\noindent \textbf{RQ4 – Effect of $N$ on ORM Performance.}
\emph{How the number of candidates $N$ influence the performance of ORM-based selection?}
Table~\ref{tab:n_decreasing} present the effect of varying $N$ on execution accuracy across \textit{BIRD dev}, \textit{Spider dev}, and \textit{Spider test}. We compare four strategies: execution-based Best-of-$N$, Majority Voting, ORM-based Best-of-$N$, and Pass@$N$, the latter serving as an upper bound on achievable performance by reporting the proportion of queries for which at least one correct candidate is generated. The results show that ORM-based selection consistently achieves higher accuracy than heuristic methods across all values of $N$, while also approaching the upper bound defined by Pass@$N$ more closely as $N$ increases.

On \textit{BIRD dev}, ORM accuracy increases from 63.89\% at $N=1$ to 68.71\% at $N=25$, after which it stabilizes with only minor fluctuations, reaching 68.90\% at $N=32$. In contrast, execution-based Best-of-$N$ plateaus earlier around 66.0\%, while Majority Voting converges near 67.0\%. A similar pattern emerges on \textit{Spider dev}, where ORM steadily improves to 84.62--84.72\% in the range $N=17$--$30$, compared to 82.88\% for execution-based Best-of-$N$ and 84.14\% for Majority Voting. On \textit{Spider test}, ORM achieves 87.47\% at $N=32$, maintaining a consistent margin of +2 to +3 points over both baselines. 
Overall, these results show that ORMs not only scale more effectively with increasing candidate pools but also sustain their advantage as $N$ grows, underscoring their effectiveness as a selection mechanism.

Figure~\ref{fig:n_decreasing_difficulty} further decomposes performance by query difficulty (\textit{Simple}, \textit{Moderate}, \textit{Challenging}) on the \textit{BIRD} dev set. For simple queries, all methods improve with increasing $N$, but ORM consistently maintains a margin of 1--2 points over Majority Voting and 2--3 points over execution-based Best-of-$N$. The advantage becomes more pronounced for moderate queries, where ORM shows a smooth upward trajectory and stabilizes near 63\%, while both baselines plateau earlier at lower levels. The most striking gains are observed on challenging queries: here, ORM substantially outperforms both strategies, with improvements exceeding 3--4 points at higher $N$, whereas execution-based Best-of-$N$ remains flat and Majority Voting oscillates with unstable progress. 
These results highlight that ORM selection is particularly effective in harder scenarios, where discriminating between semantically correct and incorrect candidates becomes critical.

Overall, these findings show that ORM-based selection benefits consistently from larger candidate pools, outperforming the other heuristic strategies across datasets, evaluation splits, and difficulty levels. While Majority Voting and execution-based Best-of-$N$ provide incremental improvements at low values of $N$, they quickly saturate, whereas ORM continues to scale and more closely approaches the oracle bound defined by Pass@$N$. This reinforces the conclusion that ORMs provide a principled and effective mechanism for leveraging increased generation budgets, ensuring more reliable gains from test-time compute.

\begin{table}[!t]
\centering
\caption{Execution accuracy (\%) and Pass@32 on the \textit{BIRD dev}, \textit{Spider dev}, and \textit{Spider test}. Candidate queries are generated by OmniSQL models with 7B, 14B, and 32B parameters, while the ORM is implemented using OmniSQL-7B. Best scores per column are shown in \textbf{bold}.}
\label{tab:generator_comparison}


\resizebox{\columnwidth}{!}{%
  \begin{tabular}{c
                  l
                  S[table-format=2.2]
                  S[table-format=2.2]
                  S[table-format=2.2]
                  S[table-format=2.2]
                  S[table-format=2.2]}
    \toprule
    & \textbf{Model} & {Baseline} & {Ex-BoN} & {Majority Voting} & {ORM-BoN} & {Pass@32} \\
    \midrule
    \multirow{3}{*}{\rotatebox{90}{\parbox{0.8cm}{\centering \scriptsize \textbf{BIRD dev}}}} 
      & OmniSQL-7B  & 63.89 & 66.04 & 66.95 & 68.90 & 80.57 \\
      & OmniSQL-14B & \bfseries 64.28 & \bfseries 66.75 & \bfseries 67.01 & 69.04 & 80.90 \\
      & OmniSQL-32B & 64.02 & 66.10 & 63.95 & \bfseries 69.30 & \bfseries 82.20 \\
    \midrule
    \multirow{3}{*}{\rotatebox{90}{\parbox{0.8cm}{\centering \scriptsize \textbf{Spider dev}}}} 
      & OmniSQL-7B  & 82.40 & 82.79 & \bfseries 83.75 & 84.33 & 91.68 \\
      & OmniSQL-14B & 82.01 & 82.20 & 83.56 & \bfseries 84.62 & 92.75 \\
      & OmniSQL-32B & \bfseries 83.17 & \bfseries 83.37 & 83.08 & 84.43 & \bfseries 92.84 \\
    \midrule
    \multirow{3}{*}{\rotatebox{90}{\parbox{0.8cm}{\centering \scriptsize \textbf{Spider test}}}} 
      & OmniSQL-7B  & 84.02 & 85.14 & 85.47 & 87.33 & 93.29 \\
      & OmniSQL-14B & \bfseries 84.91 & \bfseries 85.28 & 84.86 & \bfseries 87.42 & 93.53 \\
      & OmniSQL-32B & 84.30 & 84.72 & \bfseries 86.07 & 87.19 & \bfseries 94.08 \\
    \bottomrule
  \end{tabular}
}
\end{table}
\noindent \textbf{RQ5 – ORM Effectiveness with Larger Generator Models.}
\emph{How does ORM performance vary when the underlying generator model is scaled to larger parameter sizes?}
Table~\ref{tab:generator_comparison} reports execution accuracy and Pass@32 across OmniSQL models of different sizes (7B, 14B, 32B) on \textit{BIRD dev}, \textit{Spider dev}, and \textit{Spider test}. The ORM is fixed to OmniSQL-7B, while the generator varies in scale. Results show that ORM-based selection remains consistently competitive across all generator sizes, outperforming Majority Voting (Maj) and Execution-based Best-of-$N$ (ex-BoN) methods, while approaching the oracle-like upper bound defined by Pass@32.

On \textit{BIRD dev}, ORM achieves the highest accuracy with all generator sizes, reaching 68.90\% with 7B, 69.04\% with 14B, and 69.30\% with 32B, demonstrating that ORM continues to provide robust gains even as the generator becomes larger. Interestingly, while Majority Voting and Best-of-$N$ improve moderately with 14B, they fall behind at 32B, where ORM remains stable and sustains its advantage. On \textit{Spider dev}, ORM also achieves the strongest results, with 84.33\% (7B), 84.62\% (14B), and 84.43\% (32B), consistently outperforming heuristic baselines by 0.5–2 points. A similar trend is observed on \textit{Spider test}, where ORM yields 87.33\% (7B), 87.42\% (14B), and 87.19\% (32B), surpassing both Maj and ex-BoN in nearly all settings.

These results suggest that ORM effectiveness is not diminished by scaling the generator. While larger models naturally improve baseline performance, the relative improvements provided by ORM remain intact. Moreover, ORM demonstrates greater stability across datasets and evaluation splits, whereas other heuristics exhibit fluctuations depending on model size. The consistently strong performance of ORM with larger generators indicates that its benefits are orthogonal to scale: ORM adds value not by replacing the advantages of larger models, but by maximizing the utility of the generated candidate set regardless of size.

In summary, ORM-based selection delivers consistent accuracy improvements even as the underlying generator grows to 14B and 32B parameters. By sustaining its advantage across scales and datasets, ORM establishes itself as a robust and scalable strategy for test-time SQL generation, ensuring reliable gains whether deployed with medium or large generator backbones.





\begin{table}[t]
\centering
\caption{Ablation study on verification prompt design for ORM on \textit{BIRD dev}. Execution accuracy (\%) by query difficulty. Generator: OmniSQL-7B; ORM: Qwen2.5-7B-Instruct. Best scores in \textbf{bold}.}
\label{tab:ablation_prompt}

\footnotesize

\resizebox{\columnwidth}{!}{%
    \begin{tabular}{lcccc}
        \toprule
        \textbf{Prompt Variant} & \textbf{Simple} & \textbf{Moderate} & \textbf{Challenging} & \textbf{Total} \\
        \midrule
        Instruction & 72.65 & 59.70 & 51.32 & 66.72 \\ 
        Data-Only & 72.76 & 59.48 & 51.72 & 66.75 \\ 
        SQL-Only & 72.86 & \textbf{63.36} & \textbf{53.79} & \textbf{68.19} \\
        Data + SQL & \textbf{73.08} & 62.50 & 50.34 & 67.73 \\
        \bottomrule
    \end{tabular}
}
\end{table}

\noindent \textbf{RQ6 – Impact of Training Prompt on ORM Results.} 
\textit{How does the design of verification prompts affect ORM performance?}
Table~\ref{tab:ablation_prompt} presents an ablation study on prompt design for ORM verification using Qwen2.5-7B-Instruct as the verifier and OmniSQL-7B as the generator, evaluated on the \textit{BIRD dev} set. Four prompt variants are compared: (i) \textit{Instruction}, which provides the SQL query with explicit verification instructions (Appendix~\ref{prompt:instruct_prompt_orm}); (ii) \textit{Data-Only}, which provides only the result set produced by executing the query (Appendix~\ref{prompt:result-set_only}); (iii) \textit{SQL-only}, which includes only the query (Appendix~\ref{prompt:sql-only_orm}); and (iv) \textit{Data+SQL}, which provides both the query and its result set (Appendix~\ref{prompt:result-set_and_sql_orm}). Results show that the choice of prompt can influence ORM effectiveness, particularly for moderate and challenging queries, while differences are less pronounced for simple ones.

For simple queries, all prompts yield nearly identical results, with accuracies clustering around 72–73\%, suggesting that prompt design plays a minimal role when the task is straightforward and candidates are easy to discriminate. Differences emerge more clearly for moderate queries, where the SQL-only prompt achieves the strongest accuracy (63.36\%), outperforming the next-best SQL+Data prompt by nearly one point, while the instruction-style and data-only variants lag behind at 59–60\%. This trend becomes even more evident in challenging queries, where the SQL-only prompt again delivers the best results (53.79\%), surpassing all other formulations by more than two points. These gains translate into the highest overall average accuracy for the SQL-only prompt (68.19\%), followed by the SQL+Data variant (67.73\%), while the instruction and data-only variants converge at roughly 66.7\%.

These findings suggest that the most effective verification signal for ORM training comes from exposing the verifier directly to SQL structures, without additional natural language instructions or contextual data. While including schema or data information can slightly improve performance in moderate cases, excessive context appears to reduce the verifier’s discriminative ability.

Overall, this ablation highlights that ORM performance is sensitive to prompt design, but the most notable gains occur in moderate and challenging queries where discrimination is more difficult. The SQL-only prompt emerges as the most effective design, establishing that lightweight, structure-focused inputs are more beneficial for ORM training than verbose instruction-style formulations.

\begin{table}[t]
\centering
\caption{Impact of ORM model size with the generator fixed to OmniSQL-7B. Execution Accuracy (EX, \%) and Pass@32 are reported on \textit{BIRD dev}, \textit{Spider dev}, and \textit{Spider test}. Best results per metric and dataset are shown in \textbf{bold}.}
\label{tab:orm_size_impact}

\resizebox{0.98\columnwidth}{!}{%
    \begin{tabular}{l
                    S[table-format=2.2]
                    S[table-format=2.2]
                    S[table-format=2.2]
                    S[table-format=2.2]
                    S[table-format=2.2]
                    S[table-format=2.2]}
    \toprule
    \multirow{2}{*}{\textbf{ORM}} &
    \multicolumn{2}{c}{\textbf{BIRD dev}} &
    \multicolumn{2}{c}{\textbf{Spider dev}} &
    \multicolumn{2}{c}{\textbf{Spider test}} \\
    \cmidrule(lr){2-3}\cmidrule(lr){4-5}\cmidrule(lr){6-7}
    & \textbf{EX} & \textbf{Pass@32} & \textbf{EX} & \textbf{Pass@32} & \textbf{EX} & \textbf{Pass@32} \\
    \midrule
    OmniSQL-7B  & 68.90 & 80.57 & 84.53 & 91.68 & 87.47 & 93.29 \\
    OmniSQL-14B & \textbf{69.23} & 80.57 & \textbf{85.01} & 91.68 & \textbf{87.84} & 93.29 \\
    OmniSQL-32B & 68.19 & 80.57 & 84.23 & 91.68 & 86.91 & 93.29 \\
    \bottomrule
    \end{tabular}
}
\end{table}

\noindent \textbf{RQ7 – Effect of ORM Size Beyond 7B Models.}  
\emph{Does scaling the ORM verifier beyond 7B parameters lead to performance improvements?} Table~\ref{tab:orm_size_impact} evaluates ORM models of different sizes (7B, 14B, 32B) with the generator fixed to OmniSQL-7B. We report execution accuracy (EX) and Pass@32 on \textit{BIRD dev}, \textit{Spider dev}, and \textit{Spider test}. Results show that increasing ORM size beyond 7B yields only marginal and inconsistent improvements, suggesting diminishing returns when scaling the verifier model.

On \textit{BIRD dev}, the 14B ORM achieves the highest accuracy (69.23\%), a modest gain over the 7B variant (68.90\%), while the 32B model slightly underperforms at 68.19\%. On \textit{Spider dev}, the 14B ORM again leads with 85.01\%, compared to 84.53\% for 7B and 84.23\% for 32B. A similar pattern is observed on \textit{Spider test}, where the 14B ORM achieves 87.84\%, outperforming both 7B (87.47\%) and 32B (86.91\%). 

These findings suggest that scaling the ORM beyond 7B parameters does not yield consistent accuracy gains. The modest improvements observed with the 14B model are offset by performance drops with the 32B variant, pointing to a saturation effect where additional verifier capacity brings limited benefit. Importantly, the stability of Pass@32 across ORM sizes underscores that larger verifiers cannot compensate for the inherent limitations of the generator. In practice, this indicates that a 7B ORM is sufficient to achieve strong results, offering a favorable balance between performance and computational efficiency.

Overall, RQ7 highlights that, unlike generator scaling which provides steady improvements (RQ5), ORM scaling exhibits diminishing returns. This reinforces the idea that ORM effectiveness is primarily driven by its training signal and alignment objective rather than raw parameter count, and that small to medium-sized verifiers are adequate for practical deployment.

\begin{table}[t]
\centering
\caption{Ablation study comparing autoregressive and Binary Cross-Entropy (BCE) loss fine-tuning on \textit{BIRD dev} across query difficulty levels. Both the generator and ORM model are OmniSQL-7B. Reported metric is Execution Accuracy (\%). Best results are in \textbf{bold}.}
\label{tab:finetune_ablation}
\sisetup{
  table-format=2.2,
  output-decimal-marker = {,} 
}
\renewcommand{\arraystretch}{1.08}

\resizebox{\columnwidth}{!}{%
    \begin{tabular}{l S S S S}
    \toprule
    \textbf{Method} & \textbf{Simple} & \textbf{Moderate} & \textbf{Challenging} & \textbf{Total} \\
    \midrule
    Majority Voting & 72.86 & 59.70 & 52.41 & 66.95 \\
    BCE FT   & 72.86 & 59.91 & 51.72 & 66.95 \\
    Autoregressive FT    & \textbf{74.15} & \textbf{62.93} & \textbf{54.48} & \textbf{68.90} \\
    \bottomrule
    \end{tabular}
}
\end{table}

\noindent \textbf{RQ8 – Autoregressive vs. Cross-Entropy Training.}
\emph{How do different fine-tuning objectives influence ORM effectiveness?}
Table~\ref{tab:finetune_ablation} presents an ablation study on the \textit{BIRD dev}, comparing two fine-tuning strategies, autoregressive and Binary Cross-Entropy (BCE), against a Majority Voting baseline. In the autoregressive setting, the ORM is trained under the causal language modeling objective, where the model learns to generate the target label token (e.g., ``Yes'' or ``No'') conditioned on the input prompt. In contrast, BCE fine-tuning reformulates the task as a binary classification problem: the prompt is encoded and a classification head directly predicts the probability of correctness. Both the generator and ORM are instantiated from OmniSQL-7B, and we report execution accuracy across queries of varying difficulty levels. Results show that autoregressive fine-tuning consistently outperforms BCE training and the Majority Voting baseline, yielding the strongest overall performance.

For simple queries, all three methods perform comparably, with accuracies around 72–74\%. The autoregressive ORM reaches 74.15\%, slightly outperforming BCE (72.86\%) and Majority Voting (72.86\%). The differences become more apparent in moderate queries, where autoregressive fine-tuning achieves 62.93\%, compared to 59.91\% for BCE and 59.70\% for Majority Voting. On challenging queries, the advantage is again clear: the autoregressive ORM reaches 54.48\%, while BCE and Majority Voting lag behind at 51.72\% and 52.41\%, respectively. These improvements accumulate to the highest total accuracy of 68.90\% for the autoregressive ORM, surpassing both BCE (66.95\%) and Majority Voting (66.95\%).


These findings indicate that training ORMs with an autoregressive objective improves performance across all levels, with the largest gains in harder queries. In contrast, BCE fine-tuning, appears less effective at handling moderate and challenging cases.

Overall, with RQ8 we demonstrates that the choice of training objective has a significant impact on ORM effectiveness. While losses such as BCE can match heuristics like Majority Voting, autoregressive fine-tuning consistently delivers superior results, establishing it as the most effective approach for aligning ORMs with the requirements of the Text-to-SQL verification task.

\section{Conclusions}
We introduced \textit{GradeSQL}, a framework for synthesizing data and training Outcome Reward Models (ORMs) for the Text-to-SQL task, enabling the ranking of SQL queries generated by LLMs. \textit{GradeSQL} demonstrates how ORMs can be leveraged as an effective heuristic strategy for Best-of-$N$ (BoN) and as an alternative Test-Time Inference (TTI) approach to execution-based BoN (ex-BoN) and Majority Voting (Maj).

Our study shows that ORMs can be consistently trained across heterogeneous model families, benefit from low-scale yet balanced datasets, and achieve competitive performance on the BIRD and Spider benchmarks compared to ex-BoN and Maj for test-time inference. Moreover, ORM effectiveness remains robust across different values of $N$, scales with stronger generator models, and is influenced by design choices such as training objectives and prompt formulation.

These findings indicate that ORMs provide an effective and semantically aligned mechanism for candidate selection in Text-to-SQL. We also observe diminishing returns when increasing ORM size, suggesting that lightweight verifiers are sufficient for practical deployment.

In conclusion, \textit{GradeSQL} offers the community a unified framework for building ORMs in the Text-to-SQL setting, and our results underscore the potential of ORM-based selection as a competitive alternative to classical test-time inference strategies. Furthermore, we are releasing the framework\footnote{\href{https://github.com/sisinflab/GradeSQL}{GradeSQL Framework}}, datasets\footnote{\href{https://huggingface.co/collections/sisinflab-ai/gradesql-training-datasets-68ac62e1356b5399ef81236c}{GradeSQL Training Datasets}}, and pretrained ORMs\footnote{\href{https://huggingface.co/collections/sisinflab-ai/gradesql-models-68ac58755ffe5fe880e0acb5}{Pretrained GradeSQL ORMs}} as open-source resources to support the community and foster further research in this direction.

\appendix
\section{Prompts used}
\subsection{Zero-Shot CoT SQL Generation Prompt Template}\label{prompt:SQL_candidate_generation_prompt}
\begin{tcolorbox}[
  enhanced,
  breakable,
  width=0.98\columnwidth,   
  colback=myBlue!5!white,
  colframe=myBlue!50!black,
  boxsep=3pt,
  left=6pt,right=6pt,top=6pt,bottom=6pt,
  fontupper=\small           
]
\phantomsection

\textbf{System:}  
\textbf{"}You are a data science expert. Below, you are provided with a database schema and a natural language question. Your task is to understand the schema and generate a valid SQL query to answer the question.

\medskip

Database Engine: \texttt{SQLite}

\medskip

Database Schema: \{\texttt{db\_details}\}\\ 
This schema describes the database’s structure, including tables, columns, primary keys, foreign keys, and any relevant relationships or constraints.

\medskip

Question: \{\texttt{question}\}

\medskip

Instructions:\\
- Make sure you only output the information that is asked in the question. If the question asks for a specific column, make sure to only include that column in the SELECT clause, nothing more.\\  
- The generated query should return all of the information asked in the question without any missing or extra information. \\
- Before generating the final SQL query, please think through the steps of how to write the query.

\medskip

Output Format:\\
In your answer, please enclose the generated SQL query in a code block:
\begin{verbatim}
```
Your SQL query
```
\end{verbatim}

\medskip

Take a deep breath and think step by step to find the correct SQL query.\textbf{"}

\end{tcolorbox}

\subsection{SQL-Only Verification Prompt for ORM Evaluation}\label{prompt:sql-only_orm}
\begin{tcolorbox}[
  enhanced,
  breakable,
  width=0.98\columnwidth,   
  colback=myBlue!5!white,
  colframe=myBlue!50!black,
  boxsep=3pt,
  left=6pt,right=6pt,top=6pt,bottom=6pt,
  fontupper=\small           
]
\phantomsection

\small{
\textbf{System:} \textbf{"}Question: \texttt{\{db\_schema + natural language question\}} \\
SQL: \texttt{\{sql candidate\}} \\
Is the SQL correct?\textbf{"}
}
\end{tcolorbox}


\subsection{Data-Only Verification Prompt for ORM Evaluation}\label{prompt:result-set_only}
\begin{tcolorbox}[
  enhanced,
  breakable,
  width=0.98\columnwidth,   
  colback=myBlue!5!white,
  colframe=myBlue!50!black,
  boxsep=3pt,
  left=6pt,right=6pt,top=6pt,bottom=6pt,
  fontupper=\small           
]
\phantomsection

\small{
\textbf{System:} \textbf{"}Question: \texttt{\{db\_schema + natural language question\}} \\
Data retrieved: \texttt{\{data\}} \\
Based on data and question, in your opinion is the SQL correct?\textbf{"}
}
\end{tcolorbox}

\subsection{Data + SQL Verification Prompt for ORM Evaluation}\label{prompt:result-set_and_sql_orm}
\begin{tcolorbox}[
  enhanced,
  breakable,
  width=0.98\columnwidth,   
  colback=myBlue!5!white,
  colframe=myBlue!50!black,
  boxsep=3pt,
  left=6pt,right=6pt,top=6pt,bottom=6pt,
  fontupper=\small           
]
\phantomsection

\small{
\textbf{System:} \textbf{"}Question: \texttt{\{db\_schema + natural language question\}} \\
Data: \texttt{\{data\}} \\
SQL: \texttt{\{sql candidate\}} \\
Is the SQL correct?\textbf{"}
}
\end{tcolorbox}

\subsection{Instruction Verification Prompt for ORM Evaluation}\label{prompt:instruct_prompt_orm}
\begin{tcolorbox}[
  enhanced,
  breakable,
  width=0.98\columnwidth,   
  colback=myBlue!5!white,
  colframe=myBlue!50!black,
  boxsep=3pt,
  left=6pt,right=6pt,top=6pt,bottom=6pt,
  fontupper=\small           
]
\phantomsection

\small{
    \textbf{System:} \textbf{"}You are an expert SQL evaluator. Given a database schema, a question, the SQL query and data returned, determine if the SQL query correctly answers the question based on the schema. Respond only with Yes or No.\\
    Input:
    
    \medskip
    
    Question and database schema: \texttt{\{db\_schema + natural language question\}}\\
    - SQL Query: \texttt{\{sql candidate\}}\\
    - Data returned: \texttt{\{data\}}
    
    \medskip
    
    Instructions:\\
    1. Analyze the database schema to understand the table structure, columns, and relationships.\\
    2. Evaluate the provided SQL query against the question to determine if it accurately retrieves the intended data.\\
    3. Respond solely with Yes if the SQL query is correct, or No if it is incorrect.\\
    Output:\\
    Yes\\
    or\\
    No\\
    Is the SQL correct?\textbf{"}
}
\end{tcolorbox}



\subsection{Inference schema}
\begin{figure}[h!]
    \centering
    \resizebox{\columnwidth}{!}{%
        \includegraphics{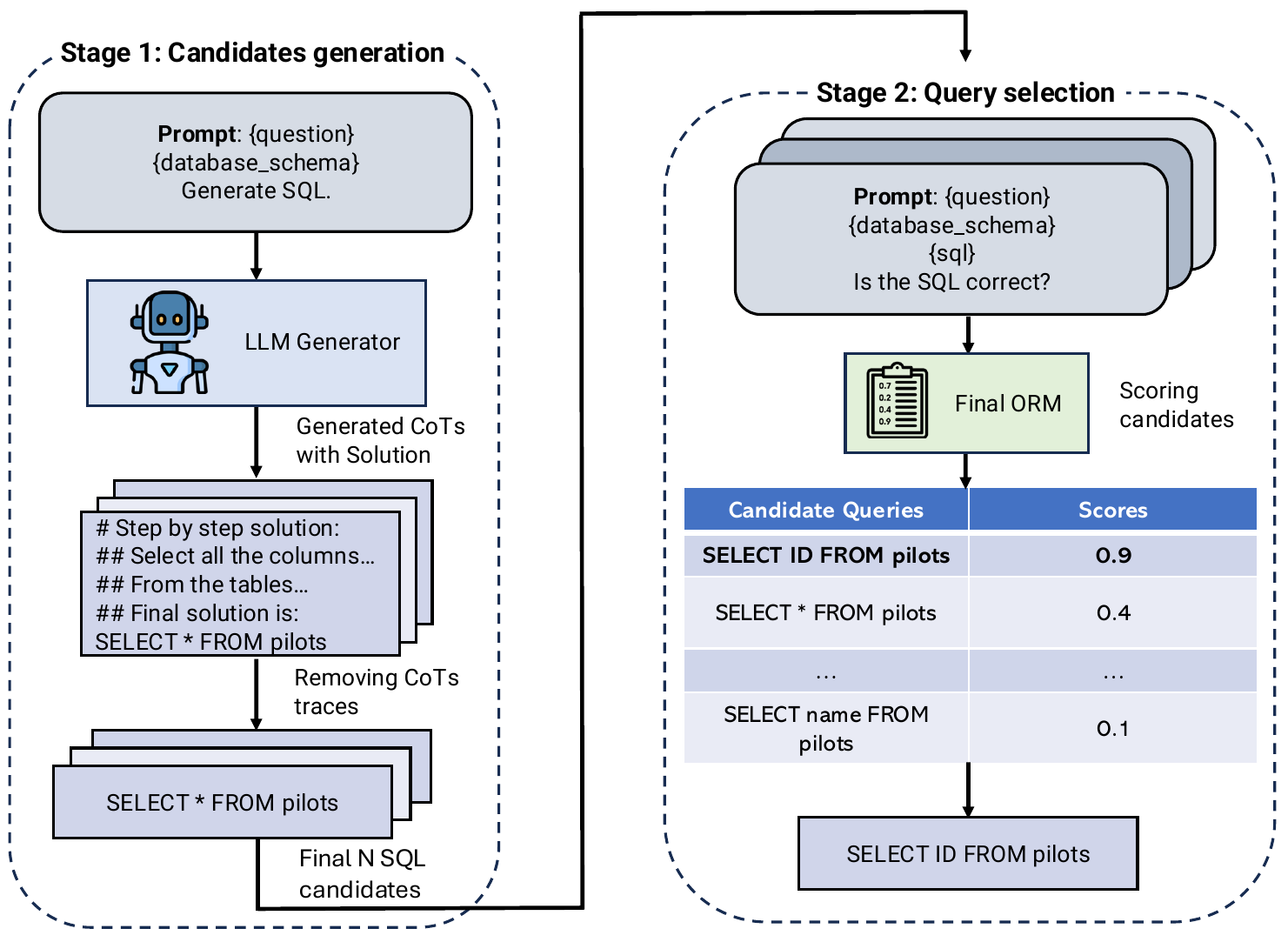}
    }
    \caption{Overview of the GradeSQL ORM at inference. The pipeline consists of two stages: (i) Candidate Generation, where an LLM produces a pool of $N$ SQL candidates; and (ii) Query Selection, where the ORM scores each candidate and ranks them, selecting the query with the highest probability as the final output.}
    \label{fig:inference_schema}
\end{figure}

\section*{Acknowledgment of AI-Generated Content}
This paper used a generative AI tool (ChatGPT) for language refinement and LaTeX formatting assistance only. All ideas, analysis, and results are original.



\end{document}